\documentclass[10pt,twocolumn,letterpaper]{article}

\usepackage[pagenumbers]{cvpr} 

\usepackage[utf8]{inputenc} 
\usepackage[T1]{fontenc}    
\usepackage{url}            
\usepackage{booktabs}       
\usepackage{amsfonts}       
\usepackage{nicefrac}       
\usepackage{microtype}      
\usepackage{bbm}
\usepackage{amsmath}
\usepackage{amssymb}
\usepackage{bbding}
\usepackage{algorithm}
\usepackage{listings}
\usepackage{multirow}
\usepackage{wrapfig}
\usepackage[dvipsnames,table,xcdraw]{xcolor}
\usepackage{graphicx}
\usepackage{caption}
\usepackage{subcaption}
\usepackage{pifont}
\usepackage{makecell}
\newcommand{\cmark}{\ding{51}}%
\newcommand{\xmark}{\ding{56}}

\newcommand{\figref}[1]{Fig.~\ref{#1}}
\newcommand{\tabref}[1]{Tab.~\ref{#1}}
\newcommand{\eqnref}[1]{Eqn.~(\ref{#1})}

\newcommand{\secref}[1]{Sec.~\ref{#1}}
\newcommand{\myPara}[1]{\vspace{6pt}\noindent\textbf{#1}}

\newcommand{\highlight}[1]{\textbf{\textcolor{ForestGreen}{#1}}}

\newcommand{\addFig}[1]{}
\newcommand{\addFigs}[1]{}
\definecolor{cvprblue}{rgb}{0.21,0.49,0.74}
\usepackage[pagebackref,breaklinks,colorlinks,linkcolor=BrickRed,citecolor=cvprblue]{hyperref}

\def\MyMthd{CorrMatch}
\def\etal{\emph{et al.\ }}

\newcommand{\tablestyle}[2]{\setlength{\tabcolsep}{#1}\renewcommand{\arraystretch}{#2}\centering\small}

\def\paperID{541} 
\def\confName{CVPR}
\def\confYear{2024}

\graphicspath{{./figures/}}

\begin{document}
\title{\MyMthd{}: Label Propagation via Correlation Matching for Semi-Supervised Semantic Segmentation}

\author{Boyuan Sun$^1$ \quad Yu-Qi Yang$^1$ \quad Le Zhang$^2$ \quad Ming-Ming Cheng$^1$ \quad Qibin Hou$^1$\thanks{Corresponding author.} \\ \\
$^1$VCIP, School of Computer Science, Nankai University \\
$^2$School of Information and Communication Engineering, UESTC
}
\maketitle

\begin{abstract}

This paper presents a simple but performant semi-supervised semantic segmentation approach, called \MyMthd{}.
Previous approaches mostly employ complicated training strategies to leverage unlabeled data but overlook the role of correlation maps in modeling the relationships between pairs of locations.
We observe that the correlation maps not only enable clustering pixels of the same category easily but also contain good shape information, which previous works have omitted.
Motivated by these, we aim to improve the use efficiency of unlabeled data by designing two novel label propagation strategies.
First, we propose to conduct pixel propagation by modeling the pairwise similarities of pixels to spread the high-confidence pixels and dig out more.
Then, we perform region propagation to enhance the pseudo labels with accurate class-agnostic masks extracted from the correlation maps.
\MyMthd{} achieves great performance on popular segmentation benchmarks.
Taking the DeepLabV3+ with ResNet-101 backbone as our segmentation model, we receive a 76\%+ mIoU score on the Pascal VOC 2012 dataset with only 92 annotated images.
Code is available at \href{https://github.com/BBBBchan/CorrMatch}{https://github.com/BBBBchan/CorrMatch}.
\end{abstract}
\vspace{-10pt}
\section{Introduction}
\label{sec:intro}

With the development of deep learning techniques, especially convolutional neural networks (CNNs)~\cite{he2016deep, pami21Res2net, xie2017aggregated, zhang2022resnest, 22PAMI-RF-Next}, many significant semantic segmentation methods~\cite{long2015fully,zhao2017pyramid, chen2018encoder, guo2022segnext, ronneberger2015u} have achieved remarkable results.
However, methods based on deep learning often require large-scale pixel-wise annotated datasets with a massive amount of labeled images. 
Compared to the image classification and object detection tasks~\cite{deng2009imagenet,lin2014microsoft}, the accurate annotations for segmentation datasets are very expensive and time-consuming.

Recently, many researchers have sought to address the above challenge by reducing the demand for large-scale accurately annotated data in the semantic segmentation task by presenting weakly-supervised~\cite{wei2018revisiting,21TPAMI-OAA, wang2020self, Jiang2022L2G}, semi-supervised~\cite{hong2015decoupled,hu2021semi,french2020semi, ouali2020semi}, or even unsupervised segmentation methods~\cite{hwang2019segsort, Gao22LUSS,van2021unsupervised, harb2022infoseg}.
Among these schemes, semi-supervised semantic segmentation only requires a small amount of labeled data accompanied by a large amount of unlabeled data for training, which approaches real-world scenarios more and hence attracts the favor of more and more researchers from academia and industry. 

\begin{figure}[t]
  \small
  \centering
  \setlength{\abovecaptionskip}{2pt}
  \includegraphics[width=0.95\linewidth]{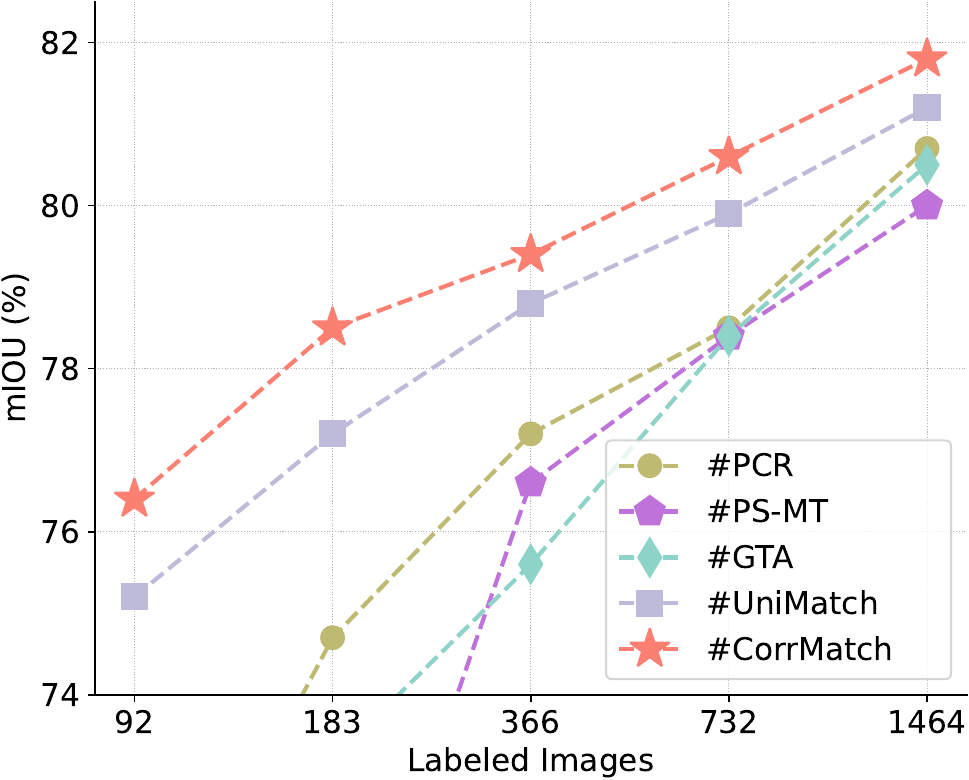}
  \put(-42, 71){\cite{xu2022semi}}
  \put(-34, 60){\cite{liu2022perturbed}}
  \put(-42, 49){\cite{jin2022semi}}
  \put(-22, 38){\cite{yang2022revisiting}}
  \caption{Comparison with state-of-the-art methods on the Pascal VOC dataset. 
    Our \MyMthd{} outperforms all others for all splits.
  }\label{fig:method_miou}
  \vspace{-20pt}
\end{figure}
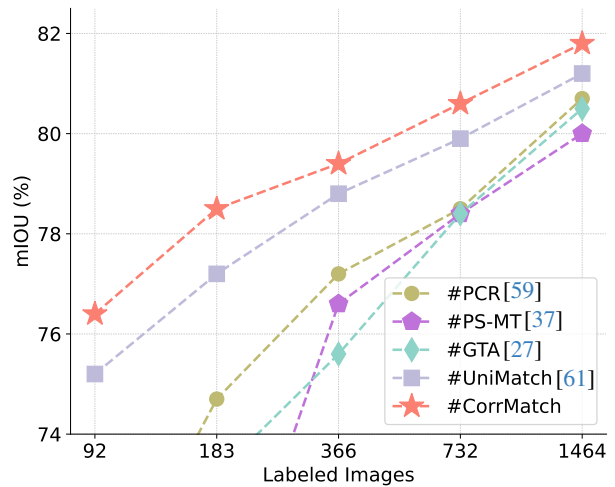

In the literature of semi-supervised semantic segmentation, most works adopt the Mean Teacher architecture~\cite{hu2021semi, liu2022perturbed, xu2022semi, jin2022semi} or self-training strategy~\cite{ke2022three,yang2022st++,yuan2021simple} to enable consistency regularization. 
As shown in \tabref{method compare}, these methods often require extra networks or training stages, complicating the training process.
Although the recent UniMatch~\cite{yang2022revisiting} has shown that a single-stage pipeline is sufficient, it still demands multiple strong augmentation data streams. 
Unlike them, our \MyMthd{} is a simpler framework with no need for multiple networks, training stages, or strong augmentation data streams.

Furthermore, in previous works~\cite{yang2022st++, liu2022perturbed, xu2022semi}, the most popular way to leverage unlabeled data is setting a fixed threshold to screen reliable pixels as pseudo labels. However, those methods often struggle to efficiently utilize unlabeled data due to the trade-off between pseudo-label proportion and accuracy via threshold adjustments.
Beyond that, motivated by the fact that the correlations between pixels can reflect the pairwise similarities, which indicates semantically similar pixels exhibit higher similarity on the correlation map, we reconsider the challenge of accurately assigning pseudo labels to unlabeled data from a label propagation perspective.

First, considering the correlation maps embed the global pairwise similarities, we propose the pixel propagation strategy. 
With correlation maps constructed from extracted features, the pixel propagation strategy spreads them into predictions, which enriches predictions with global similarities information and fosters semantic consistency. 
Meanwhile, with the observation that every row of a correlation map is equipped with local shape information, a series of binary maps that capture the objects' shapes can be acquired.
Thus, coupled with the most salient predicted class within the intersection of the shapes and high-confidence regions, we propose the region propagation strategy to enhance pseudo labels by accurately assigning class labels to these shapes.
By considering the union of shapes and high-confidence regions as the new ones, the high-confidence regions can be expanded, consequently improving the use efficiency of unlabeled data. 
As shown in \figref{fig:method_miou}, our \MyMthd{} outperforms all previous approaches.


\begin{table}[t!]
  \centering
  \small
  \setlength{\abovecaptionskip}{2pt}
  \setlength{\tabcolsep}{2pt}
  \caption{Differences between our \MyMthd{} and some representative approaches. 
    SDA denotes strong data augmentation.
  }\label{method compare}
  \begin{tabular}{lcccc} \toprule
    \makecell{Method} & \makecell{Multiple \\ networks} &\makecell{Multi-train \\ stages} & \makecell{Multiple SDA \\ streams} & \makecell{Pairwise\\  similarity}\\ \midrule
    PS-MT~\cite{liu2022perturbed}& \cmark & \xmark & \xmark & \xmark\\
    ST++~\cite{yang2022st++} & \xmark & \cmark & \xmark & \xmark\\
    ELN~\cite{kwon2022semi}& \cmark  &\cmark &\xmark & \xmark \\
    UniMatch~\cite{yang2022revisiting}& \xmark& \xmark & \cmark & \xmark \\ \midrule
    \textbf{CorrMatch}&\highlight{\xmark} & \highlight{\xmark} & \highlight{\xmark} & \highlight{\cmark} \\ \bottomrule
  \end{tabular}  
  \vspace{-20pt}
\end{table}

Our main contributions can be summarized as follows:
\begin{itemize}
  \item We demonstrate the two advantages of correlation maps in improving the use efficiency of unlabeled data. 
  \item We design a simple but performant semi-supervised semantic segmentation framework containing two novel label propagation strategies.
  \item Our \MyMthd{} achieves new state-of-the-art performance on the Pascal VOC 2012 and Cityscapes datasets without any computation burden during inference.
\end{itemize}

\section{Related Work} \label{realted}

\subsection{Semi-Supervised Learning}

Semi-supervised learning~\cite{zhu2005semi, seeger2000learning} is proposed to settle a paradigm that how to construct models using both labeled and unlabeled data and has been studied long before the deep learning era~\cite{joachims1999transductive, bennett1998semi, belkin2006manifold}. 
And certainly, semi-supervised learning has gained more attention with advancements in deep learning and computer vision~\cite{lee2013pseudo, xie2020self, NIPS2004_96f2b50b, zoph2020rethinking, berthelot2019mixmatch}. 

%
%
Since Bachman \etal~\cite{bachman2014learning} proposed a consistency regularization-based method, many approaches, such as $\Pi$-Model~\cite{laine2016temporal, sajjadi2016regularization}, Mean Teacher~\cite{tarvainen2017mean} and Dual Student~\cite{ke2019dual} have migrated it into the semi-supervised learning field.  
%
%
%
Recently, FixMatch~\cite{sohn2020fixmatch} provides a simple weak-to-strong consistency regularization framework and serves as many other relevant methods' baseline~\cite{sohn2020simple, grollmisch2021improving, upretee2022fixmatchseg, yang2022revisiting}. 
However, many follow-up works~\cite{wang2022freematch, zhang2021flexmatch, xu2021dash} have pointed out that simply setting a manually fixed threshold may lead to inferior performance and slow convergence speed. 
Among them, FreeMatch~\cite{wang2022freematch} provides a dynamic threshold scheme connected with the model's learning process.
However, these strategies designed for classification are not suitable for segmentation as multiple categories often exist in each image.

\subsection{Semi-Supervised Semantic Segmentation}

As semi-supervised learning has achieved surprising results in the image classification~\cite{lee2013pseudo, tarvainen2017mean, sohn2020fixmatch, laine2016temporal}, many works adopt the same setting for semantic segmentation~\cite{hong2015decoupled,ouali2020semi, xiao2022semi}.

One type of methods~\cite{french2020semi, zhou2021c3, hu2021semi, wang2022semi, liu2022perturbed, xu2022semi, zhang2022region, zhen23imas} adopt the Mean Teacher architecture.
U$^2$PL~\cite{wang2022semi} attempts to use unreliable predictions via contrastive learning better.
PS-MT~\cite{liu2022perturbed} builds a stricter teacher with the VAT~\cite{miyato2018virtual} technique.
ELN~\cite{kwon2022semi} uses an error localization network to mitigate the performance degradation caused by confirmation bias due to invalid pseudo labels.
%
%
All of these methods demand multi-networks for training.
Meanwhile,  another type of method, self-training based methods~\cite{ke2022three,yang2022st++,yuan2021simple,du2022learning}, often require multiple training stages. Among them, ST++~\cite{yang2022st++} proposes a three-stage paradigm with strong augmentation.
SimpleBase~\cite{yuan2021simple} uses separated batch normalization~\cite{ioffe2015batch} for images with different augmentation. 
PC$^2$Seg~\cite{zhong2021pixel} uses feature-space contrastive learning besides consistency training. 
Recently, UniMatch~\cite{yang2022revisiting} adopted a single-stage framework based on FixMatch~\cite{sohn2020fixmatch} via multiple strong augmentation branches.
%
Unlike all the above, \MyMthd{} explores how to take advantage of correlation maps better to improve the use efficiency of unlabeled data via label propagation, which previous works have ignored.




\begin{figure*}[t]
  \centering
  \setlength{\abovecaptionskip}{2pt}
  \includegraphics[width=0.98\textwidth]{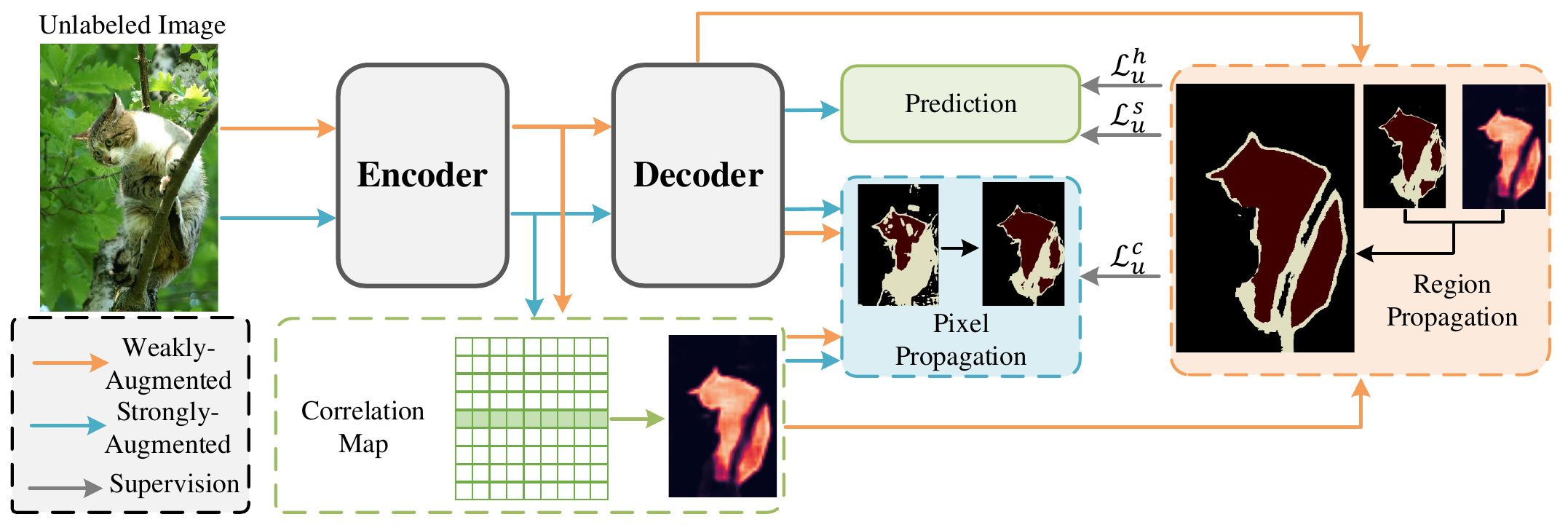}
  \caption{Illustration of our \MyMthd{} pipeline for unlabeled images. 
    We build it upon the DeepLabv3+ framework~\cite{chen2018encoder}. 
    Besides consistency regularization, \MyMthd{} adopts two label propagation strategies with correlation matching.
  }\label{fig:pipeline}
\end{figure*}

\section{\MyMthd{}}

The goal of semi-supervised semantic segmentation is to train a semantic segmentation network $\mathcal{F}$ with a small labeled image set and a large unlabeled image set.
We present a single-stage framework \MyMthd{}, which leverages pairwise correlations to achieve two label propagation strategies.

\subsection{Preliminaries} \label{framework}

\MyMthd{} is built upon a simple framework \cite{yang2022revisiting} with weak-to-strong consistency regularization. 
A standard cross-entropy loss is applied for labeled images $\{x^l_i\}$ and their corresponding labels $\{y^l_i\}$.
And unlabeled images $\{x^u_i\}$ are mainly leveraged by enforcing prediction consistency.
For an unlabeled image, $x^w_i$ and $x^s_i$ represent its augmented version with weak and strong augmentation, respectively. 
The consistency regularization treats the prediction of $x^w_i$ as the pseudo label for $x^s_i$.
We demonstrate the pipeline of unlabeled images in \figref{fig:pipeline}.

Given a mini-batch of $N$ unlabeled images, we encourage the outputs to be consistent for both weakly and strongly augmented inputs with hard supervision:
\begin{equation}\label{eqn:unlabel-hard-loss}
    \mathcal{L}_u^{h} = \frac{1}{N} \sum^{N}_i \ell_{c}(\mathcal{F}(x^s_i) , \mathcal{F}(x^w_i) ) \odot \mathcal{M}_i,
\end{equation}
where $\ell_{c}$ is the pixel-wise cross-entropy loss function and $\odot$ is the element-wise multiplication. 
$\mathcal{M}_i$ is a binary map indicating the positions with high confidence predictions in $\mathcal{F}(x^w_i)$, which can be written as:
\begin{equation}\label{eqn:filter-map}
  \mathcal{M}_i = \mathbbm{1}(\max(\hat{\mathcal{F}}(x^w_i)) > \tau),
\end{equation}
where $\hat{\mathcal{F}}(x^w_i) \in \mathbb{R}^{ K \times HW}$ is the logits output produced by the semantic segmentation network $\mathcal{F}$ and $K$ is the class number. 
$\tau$ is a threshold used to screen high-confidence predicted pixels as the pseudo label. 

However, $\mathcal{L}_u^{h}$ only treats $\mathcal{F}(x^w_i)$ as the hard pseudo label and thus ignores additional information stored in logits $\hat{\mathcal{F}}(x^w_i)$. 
Taking this into account, we further consider the consistency between the logits of the weakly and strongly augmented images in high-confidence regions.
In \eqnref{eqn:soft}, we give the formula of $\mathcal{L}_{u}^{s}$ for soft supervision.
\begin{equation}\label{eqn:soft}
  \mathcal{L}_{u}^{s} = \frac{1}{N} \sum^{N}_{i=1} \text{KL} (\hat{\mathcal{F}}(x^s_i), \hat{\mathcal{F}}(x^w_i) ) \odot \mathcal{M}_i,
\end{equation}
where $\text{KL}(\cdot) $ is Kullback-Leibler Divergence loss function. 
%
%
We view the above framework as our baseline.

\subsection{Pixel Propagation} \label{sec:correlation-maps}

As discussed in \secref{sec:intro}, pseudo labels obtained through threshold-based selection overlook the semantic similarity between pixels, constraining the utilization of unlabeled data.
In this section, we propose the pixel propagation strategy to enhance the model's overall awareness of pairwise similarities and consequently improve the utilization of unlabeled data, which involves two steps: (1) calculating correlation maps and (2) spreading correlation maps into predictions.

We first extract features $w_1$ and $w_2 \in \mathbb{R}^{D \times HW}$ through linear layers after the encoder of the network, where $D$ is the channel dimension and $HW$ is the number of feature vectors.
%
%
These extracted features enable correlation matching to quantify the degree of pairwise similarity.
Thus, we compute the correlation map $\mathcal{C}$ by performing a matrix multiplication between all pairs of feature vectors:
\begin{equation}
    \mathcal{C} = \mathrm{Softmax}({w}_1^{\top}\cdot {w}_2)/\sqrt{D},
\end{equation}
where $^\top$ denotes the matrix transpose operation. 
The correlation map $\mathcal{C} \in \mathbb{R}^{H W \times HW}$ is a 2D matrix and is activated by a $\mathrm{Softmax}$ function to yield pairwise similarities. 
$\mathcal{C}$ enables accurate delineation of the corresponding regions belonging to the same object as shown in \figref{fig:pipeline} and inspires us to propagate it into pseudo labels using correlation matching. 
More visualizations can be found in \figref{fig:enhance}.

To enhance the model's awareness of pairwise similarity, we spread the correlation map $\mathcal{C}$ into model logits outputs $\hat{\mathcal{F}}(x^{u}_{i})$ to attain another representation of the prediction $\mathbf{z}_i^u \in  \mathbb{R}^{ K \times HW}$ via label propagation:
\begin{equation}
\label{eqn:reaction}
    \mathbf{z}_i^u = f_1(\hat{\mathcal{F}}(x^{u}_{i})) \cdot  \mathcal{C},
\end{equation}
where $f_1(\cdot)$ is a bilinear interpolation for shape matching. 
The resulting $\mathbf{z}_i^u$ emphasizes the pairwise similarities of the same object through the correlation map.


%
Therefore, a correlation loss $\mathcal{L}_{u}^{c}$ can be calculated between $\mathbf{z}_i^u$ and the high-confidence pseudo labels as the supervision, which can be written as follows:
\begin{equation}
\label{eqn:correlation loss}
    \mathcal{L}_{u}^{c} = \frac{1}{|N|} \sum^{N}_{i=1} (\ell_{c}(\mathbf{z}_i^u, \mathcal{F}(x^w_i))) \odot \mathcal{M}_i.
\end{equation}
For the labeled images $\{x_i^l\}$, we also compute the cross-entropy loss between $\mathbf{z}^l_i$ and $y_i^l$ as the supervised correlation loss $\mathcal{L}_{s}^{c}$, where $\mathbf{z}^l_i$ can be attained using \eqnref{eqn:reaction}. 
So far, given a weakly augmented unlabeled image $x^w_i$, its correlation map $\mathcal{C}^{w}_i$ can effectively model pairwise similarities.

\subsection{Region Propagation}\label{sec: region prop}

\begin{figure}[t]
	\centering
	\includegraphics[width=0.98\linewidth]{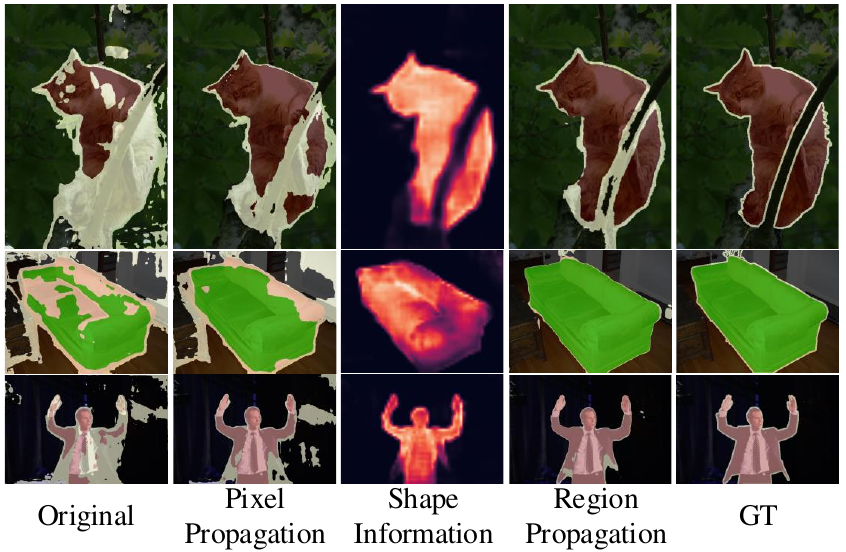}
    \setlength{\abovecaptionskip}{2pt}
	\caption{ Illustration of our proposed propagation strategies. White areas are ignored regions due to low confidence. Combining the shape information with the most salient class, \MyMthd{} can significantly enhance pseudo labels and expand high-confidence regions.
		}
	\label{fig:enhance}
\end{figure}

During experiments, we also observe that every row $\mathbf{c}$ in $\mathcal{C}^{w}_i$ denotes the similarity between individual feature vectors and all vectors within the entire feature map, which implicitly encapsulates shape information. 
With this observation, we propose the region propagation strategy to enhance pseudo labels with these shapes information. 
Specifically, we first normalize $\mathbf{c}$ and turn it into a binary map $\hat{\mathbf{c}}$:
\begin{equation}
    \hat{\mathbf{c}}= f_2(\mathbbm{1}(\frac{\mathbf{c} - \min(\mathbf{c})}{\max(\mathbf{c}) - \min(\mathbf{c})} > 0.5)),
\end{equation}
where $f_2(\cdot)$ is a shape-matching function to align the shapes of $\hat{\mathbf{c}}$ and $\mathcal{F}(x^w_i)$. 
As shown in \figref{fig:enhance}, the shape information $\hat{\mathbf{c}} \in \mathbb{R}^{H\times W}$ explicitly embeds class agnostic shape information. 
%
For every $\hat{\mathbf{c}}$, we can calculate the overlap ratio $r_1$ between $\hat{\mathbf{c}}$ and the high-confidence regions $\mathcal{M}_i$.
When $\hat{\mathbf{c}}$ has a large overlap with $\mathcal{M}_i$, (i.e., $r_1 > \tau$), we are able to use $\hat{\mathbf{c}}$ to adjust the pseudo label $\mathcal{F}(x^w_i)$.


Given the current pseudo labels $\mathcal{F}(x^w_i)$, we can calculate the quantity of each unique class $l \in L$ within high-confidence shape $( \mathcal{F}(x^w_i) \odot \mathcal{M}_i \odot \hat{\mathbf{c}} )$ by a function $G(l)$ and locate the most significant class $k^*$ with the following equation:
\begin{align} \label{find significant class}
    k^* &= \text{argmax}_{l \in L}G(l), \\
    G(l) &= \sum_{HW} \mathbbm{1}[ (\mathcal{F}(x^w_i)  \odot \mathcal{M}_i \odot \hat{\mathbf{c}}) = l],
\end{align}
where $L$ is the set of all unique classes that present in predictions $\mathcal{F}(x^w_i)$. With the most significant class $k^*$, we can calculate its proportion $r_2$ within the high-confidence shape.

When $k^*$ highly coincides with the high-confidence shape, (i.e., $r_2 > \tau$), we can propagate the specific class $k^*$ into the enhanced pseudo label $\mathcal{F}(x^w_i)$ and expanded high-confidence regions $\mathcal{M}_i$ by matching the certain shape $\hat{\mathbf{c}}$.
\begin{equation}
    \mathcal{F}(x^w_i) = \begin{cases}
                        k^*, & \hat{\mathbf{c}} = 1\\
                        \mathcal{F}(x^w_i), & \hat{\mathbf{c}} = 0\\
                        \end{cases},
    \mathcal{M}_i = \mathcal{M}_i \cup \hat{\mathbf{c}}
\end{equation}

However, considering the intricate computations required for each specific shape within the correlation map and the frequent occurrence of similar semantic information in adjacent regions, resulting in similar shapes in the correlation map, it becomes evident that involving every row of the correlation map in pseudo labels optimization is redundant. Hence, we employed a random sampling approach within the correlation map to expedite label propagation. As shown in \figref{fig:enhance}, region propagation significantly expands high-confidence regions with shape information and the most salient class.

It is also worth mentioning that the correlation map construction process and label propagation only participate in the training process and hence do not bring any additional computational burdens during the inference process.

\begin{table*}[ht]
  \tablestyle{14.5pt}{1}
  \caption{Comparisons of \MyMthd{} with the state-of-the-art approaches on the Pascal VOC 2012 val set in terms of mIoU (\%). 
    All methods are trained on the classic setting, \ie, the labeled images are selected from the original VOC train set, which consists of 1,464 images.
  }\label{tab:voc classic compare}
  \begin{tabular}{lcccccc} \toprule
    Method & Training Size & 1/16 (92) & 1/8 (183) & 1/4 (366) & 1/2 (732) & Full (1464) \\ \midrule
    ST++~\cite{yang2022st++}& 321 $\times$ 321 & 65.2 & 71.0 & 74.6 & 77.3 & 79.1 \\
    UniMatch~\cite{yang2022revisiting}& 321 $\times$ 321 & 75.2 & 77.2 & 78.8 & 79.9 & 81.2\\
    
    Mean Teacher~\cite{tarvainen2017mean}& 513 $\times$ 513 & 51.7 & 58.9 & 63.9 & 69.5 & 71.0 \\
    CutMix-Seg~\cite{french2020semi}& 513 $\times$ 513 & 52.2 & 63.5 & 69.5 & 73.7 & 76.5 \\
    PseudoSeg~\cite{zou2020pseudoseg}& 513 $\times$ 513 & 57.6 & 65.5 & 69.1 & 72.4 & 73.2 \\
    CPS~\cite{chen2021semi}& 513 $\times$ 513 & 64.1 & 67.4 & 71.7 & 75.9 & - \\
    PC $^2$ Seg~\cite{zhong2021pixel}& 513 $\times$ 513 & 57.0 & 66.3 & 69.8 & 73.1 & 74.2 \\
    U$^2$ PL ~\cite{wang2022semi}& 513 $\times$ 513& 68.0 & 69.2 & 73.7 & 76.2 & 79.5 \\
    PS-MT~\cite{liu2022perturbed}& 513 $\times$ 513& 65.8 & 69.6 & 76.6 & 78.4 & 80.0 \\
    GTA~\cite{jin2022semi}& 513 $\times$ 513& 70.0 & 73.2 & 75.6 & 78.4 & 80.5\\
    PCR~\cite{xu2022semi}& 513 $\times$ 513& 70.1 & 74.7 & 77.2 & 78.5 & 80.7 \\
    RC$^2$L~\cite{zhang2022region}&513 $\times$ 513 & 65.3 & 68.9 & 72.2 & 77.1 & 79.3 \\
    CCVC~\cite{wang2023conflict}& 513 $\times$ 513& 70.2 & 74.4 & 77.4 & 79.1 & 80.5 \\
    \rowcolor[HTML]{EFEFEF} \textbf{\MyMthd{}}& 321 $\times$ 321 & \textbf{76.4} & \textbf{78.5}  & \textbf{79.4} & \textbf{80.6} & \textbf{81.8} \\
    \bottomrule
  \end{tabular}
\end{table*}

\subsection{More Details} \label{sec: more details}

\noindent \textbf{Dynamic threshold.}
As mentioned in FreeMatch~\cite{wang2022freematch}, using a fixed threshold $\tau$ that is too strict or too loose is detrimental to model convergence. 
At the same time, we observe that the most suitable thresholds are different for different experimental settings (\figref{fig:different fix}). Thus, We provide a dynamic threshold strategy that is related to the training process.

Given the threshold $\tau$ a relatively small value (0.85) as initialization, the strategy of updating $\tau$ depends on the logits $\hat{\mathcal{F}}(x^{w}_{i}) $.
We use the exponential moving average (EMA)~\cite{polyak1992acceleration} to iteratively update $\tau$. Each increment is defined as: 
\begin{equation}
\Delta\tau = \frac{1}{|L|} \sum\limits_{l \in L} \max [ \mathbbm{1}(\mathcal{F}(x^w_i) = l) \odot \max^{c}(\hat{\mathcal{F}}(x^w_i))],
\end{equation}
where $\max\limits^{c}(\cdot)$ denotes taking the maximum value along the channel dimension. 
This operation aims to take the maximum confidence of all predicted classes in $\hat{\mathcal{F}}(x^{w}_{i}) $ and use their average as the increment for each iteration.
We found that such a simple threshold updating strategy works well. 
We will further show in \secref{ablation} that $\tau$ is insensitive to initialization.
The corresponding pseudo code is provided in the supplementary materials.

\myPara{Loss function.}
The overall objective function $\mathcal{L}$ is a combination of supervised loss $\mathcal{L}_{s}$
and unsupervised loss $\mathcal{L}_{u}$: $\mathcal{L} = \frac{1}{2}( \mathcal{L}_{s} + \mathcal{L}_{u})$.
Like most methods, we use the cross-entropy loss function $\mathcal{L}_{s}^{h}$ as the basic supervision of labeled data $\mathcal{D}^{l}$. Therefore, the supervised loss $\mathcal{L}_{s}$ is defined as the combination of $\mathcal{L}_{s}^{h}$ and supervised correlation loss $\mathcal{L}_{s}^{c}$:
$\mathcal{L}_{s} = \frac{1}{2} (\mathcal{L}_{s}^{h} + \mathcal{L}_{s}^{c})$.
As for unsupervised loss $\mathcal{L}_{u}$ on unlabeled data $\mathcal{D}^{u}$, it can be expressed as follows:
\begin{equation}
    \mathcal{L}_{u} = \lambda_1\mathcal{L}_{u}^{h} + \lambda_2\mathcal{L}_{u}^{s} + \lambda_3\mathcal{L}_{u}^{c},
\end{equation}
where $\mathcal{L}_{u}^{h}, \mathcal{L}_{u}^{s}$ and $\mathcal{L}_{u}^{c}$ denote the unsupervised hard loss, soft loss, and correlation loss. And $[ \lambda_1, \lambda_2, \lambda_3]$ are set to $[0.5, 0.25, 0.25]$ by default.

\section{Experiments}

\subsection{Experiment Setup}

\myPara{Datasets.} 
We report results on the Pascal VOC 2012 and Cityscapes datasets.
Pascal VOC 2012 is a semantic segmentation benchmark with 21 classes, consisting of 1,464 high-quality annotated images for training and 1,449 images for evaluation originally~\cite{everingham2009pascal}.  We also conduct experiments on the aug Pascal VOC 2012 dataset, which contains more coarsely annotated images from the Segmentation Boundary Dataset (SBD)~\cite{hariharan2011semantic}, resulting in 10,582 training images in total. 
Cityscapes is an urban scene understanding dataset, including 2,975 training and 500 validation images with fine annotations~\cite{cordts2016cityscapes}. It contains 19 classes of urban scenes, and all images have the resolution of 1024$\times$2048.


\myPara{Implementation details.}
Following most previous semi-supervised semantic segmentation methods, we use DeepLabV3+~\cite{chen2018encoder}  with ResNet-101~\cite{he2016deep} pre-trained on ImageNet~\cite{deng2009imagenet} as the backbone. 
For the training on the Pascal VOC 2012 dataset, we use stochastic gradient descent (SGD) optimizer with an initial learning rate of 0.001, weight decay of 1e${-4}$, crop size of 321$\times$321 or 513$\times$513, batch size of 16, and training epochs of 80.
For the Cityscapes dataset, following UniMatch~\cite{yang2022revisiting}, we use stochastic gradient descent (SGD) optimizer with an initial learning rate of 0.005, weight decay of 1e${-4}$, crop size of 801 $\times$ 801, batch size of 16, and training epochs of 240 with 4 $\times$ A40 GPUs.

As for evaluation metrics, we report the mean Intersection-over-Union (mIoU) with original images following previous papers~\cite{chen2021semi,french2020semi,liu2022perturbed} for the Pascal VOC 2012 dataset. 
For Cityscapes, same as previous methods~\cite{chen2021semi, wang2022semi, yang2022revisiting}, we apply slide window evaluation with a fixed crop in a sliding window manner and then calculate mIoU on these cropped images.
All the results are measured on the standard validation set based on single-scale inference.

\begin{table*}[ht]
  \tablestyle{8pt}{1}
  \caption{Comparisons of state-of-the-art methods on the Pascal VOC 2012 val set with mIoU (\%) metric. All methods are trained on the aug setting, i.e., the labeled images are selected from the aug VOC train set, which consists of 10, 582 images. $^{\dag}$ means using U$^2$PL~\cite{wang2022semi}'s splits.}
    \label{tab:voc aug compare}
  \begin{subtable}[t]{0.49\textwidth}
  \centering
  \renewcommand{\arraystretch}{1.0}
  \begin{tabular}{lcccc} \toprule
     \multirow{2}{*}{Method} & \multirow{2}{*}{Train size}  & 1/16  & 1/8 & 1/4  \\
    & &(662) & (1323) & (2646) \\
    \midrule
    Supervised & 321 $\times$ 321  & 65.6 & 70.4 & 72.8 \\
    ST++~\cite{yang2022st++}& 321 $\times$ 321  & 74.5 & 76.3 & 76.6  \\
    CAC~\cite{lai2021semi} & 321 $\times$ 321& 72.4 & 74.6 & 76.3 \\
    UniMatch~\cite{yang2022revisiting} & 321 $\times$ 321 & 76.5 & 77.0& 77.2\\
    \rowcolor[HTML]{EFEFEF} \textbf{\MyMthd{}}& 321 $\times$ 321& \textbf{77.6} & \textbf{77.8}  &\textbf{ 78.3}   \\
    \midrule
    U2PL$^{\dag}$~\cite{wang2022semi}& 513 $\times$ 513 & 77.2 & 79.0 & 79.3 \\
    GTA$^{\dag}$~\cite{jin2022semi}&513 $\times$ 513 & 77.8 & 80.4 & 80.5  \\
    PCR$^{\dag}$~\cite{xu2022semi}& 513 $\times$ 513& 78.6 & 80.7 & 80.7 \\
    CCVC$^{\dag}$~\cite{xu2022semi}& 513 $\times$ 513& 76.8 & 79.4 & 79.6 \\
    AugSeg$^{\dag}$~\cite{zhen23augseg} &513 $\times$ 513 & 79.3 & 81.5 & 80.5  \\
        \rowcolor[HTML]{EFEFEF} \textbf{\MyMthd{}$^{\dag}$}&513 $\times$ 513 & \textbf{81.3} & \textbf{81.9} & \textbf{80.9}  \\
    \bottomrule
    \end{tabular}
  \end{subtable}
  \hfill
  \begin{subtable}[t]{0.5\textwidth}
    \centering
    \renewcommand{\arraystretch}{1.045}
    \begin{tabular}{lcccc} \toprule
     \multirow{2}{*}{Method} & \multirow{2}{*}{Train size}  & 1/16  & 1/8 & 1/4  \\
                             &                              &(662) & (1323) & (2646) \\
    \midrule
    CutMix-Seg~\cite{french2020semi}& 513 $\times$ 513  & 71.7 & 75.5 & 77.3  \\
    CCT~\cite{ouali2020semi}& 513 $\times$ 513  & 71.9 & 73.7 & 76.5 \\ 
    GCT~\cite{ke2020guided} & 513 $\times$ 513  & 70.9 & 73.3 & 76.7  \\
    CPS~\cite{chen2021semi}& 513 $\times$ 513  & 74.5 & 76.4 &77.7 \\
    AEL~\cite{hu2021semi}& 513 $\times$ 513  & 77.2 & 77.6 & 78.1  \\
    FST~\cite{du2022learning} & 513 $\times$ 513 & 73.9 & 76.1 & 78.1\\
    ELN~\cite{kwon2022semi}& 513 $\times$ 513  & - & 75.1 & 76.6 \\
    U$^2$PL~\cite{wang2022semi}& 513 $\times$ 513  & 74.4 & 77.6 & 78.7 \\
    PS-MT~\cite{liu2022perturbed}& 513 $\times$ 513  & 75.5 & 78.2 & 78.7 \\
    AugSeg~\cite{zhen23augseg} & 513 $\times$ 513  & 77.0 & 77.3 & 78.8  \\
    \rowcolor[HTML]{EFEFEF} \textbf{\MyMthd{}}&513 $\times$ 513  &  \textbf{78.4}& \textbf{79.3} &  \textbf{79.6}  \\ 
    \bottomrule
  \end{tabular}
  \end{subtable}
\end{table*}

\subsection{Comparison with State-of-the-art Methods}

\myPara{Results on classic Pascal VOC 2012}.
We show the performance of our method with other state-of-the-art methods on the classic Pascal VOC 2012 Dataset in \tabref{tab:voc classic compare}. Our experiments are conducted on various splits of the original train set following the data partition in CPS~\cite{chen2021semi}.
%
%
On the full split, our method gets 81.8\% mIoU.
Also, \MyMthd{} achieves consistent performance gains compared to existing state-of-art approaches.
Particularly, \MyMthd{} outperforms UniMatch by 1.2\%, 1.3\%, 0.6\%, 0.7\% and 0.6\% on each split.

\myPara{Results on aug Pascal VOC 2012}. 
In  \tabref{tab:voc aug compare}, we show our performance and compare with existing methods on the aug Pascal VOC 2012 Dataset. It is clear that our results are consistently much better than the existing best ones.
Our experiments are conducted on 1/16, 1/8, and 1/4 splits, respectively. Under the 321$\times$321 training size, compared to the supervised baseline, \MyMthd{} gets +12.0\%, +7.4\%, and +5.5\% improvements. In addition, our approach outperforms UniMatch by 1.1\%, 0.8\%, and 1.1\% on each split. As for the 513$\times$513 training size, our method also consistently outperforms the current state-of-the-art methods. For instance, we get 79.3\% mIoU on the 1/8 split with a gain of around 2\% compared to AugSeg~\cite{zhen23augseg}.

We also report the results using the same splits as in U$^2$PL~\cite{wang2022semi} with 513$\times$513 training size, which contain more well-annotated labels and have higher expectations of results. Compared to the best method AugSeg~\cite{zhen23augseg}, our method gains 2.0\% improvement on the 1/16 split. Furthermore, same to other methods, we observe that, as the split size increases from 1/8 to 1/4, the performance decreases under this setting. This is because in the 1/8 split, almost all of the accurately labeled images are included, and most of the images added to the larger split are coarsely labeled, which result in no improvement in performance.

\myPara{Results on Cityscapes}. 
In \tabref{tab:cityscape}, we compare the performance of \MyMthd{} with state-of-the-art methods on the Cityscapes dataset. We follow sliding window evaluation and online hard example mining (OHEM) loss~\cite{shrivastava2016training} techniques, which have been widely applied in previous SOTA works~\cite{chen2021semi, wang2022semi, liu2022perturbed, yang2022revisiting, xu2022semi, hu2021semi}. It can be clearly seen that our method can consistently outperform other methods under all splits. 
Compared to UniMatch~\cite{yang2022revisiting}, our \MyMthd{} achieves +0.7\%, +0.6\%, +0.2\%, and +0.9\% on 1/16, 1/8, 1/4, 1/2 splits, respectively.

\subsection{Ablations Studies} \label{ablation}
In this part, we conduct a series of ablations studies to verify the designs of proposed strategies in CorrMatch. We report the results of the DeepLabV3+ network using ResNet-101 as the encoder on the original Pascal VOC 2012 dataset with training size 321 $\times$ 321.

\myPara{Effectiveness of components.} \label{sec:components} We first conduct ablation studies on different components of our \MyMthd{} to demonstrate their effectiveness in \tabref{tab:components}. With the hard unsupervised loss and dynamic threshold, we get 73.6\% on the 92 split and 80.0\% on the 1464 split.
Adding soft loss $\mathcal{L}_{u}^{s}$ as the basic framework brings 0.8\% and 0.5\% improvements. With the help of label propagation, we achieve another 2.0\% and 1.3\% improvements. These results demonstrate the effectiveness of each of our components individually. Also, replacing $\mathcal{L}_{u}^{h}$ with $\mathcal{L}_{u}^{s}$ results in a performance decrease, which illustrates the importance of $\mathcal{L}_{u}^{h}$. Finally, the complete \MyMthd{} achieves 76.4\% and 81.8\% mIoU, which is +2.8\% and +1.8\% compared to the baselines.

We also conduct experiments with the fixed threshold (0.95). It can be observed that compared to the fixed baselines (73.1\% and 79.9\%), changing it into a dynamic manner only brings +0.5\% and +0.1\%. Meanwhile, after adding all components, the corresponding improvements can be lifted to +0.9\% and +1.0\%. This proves our threshold strategy cooperates well with our label propagation strategy.

\begin{table}

    \centering
  \tablestyle{8pt}{1}
    \setlength{\tabcolsep}{4pt}
    \caption{ Comparing results of state-of-the-art algorithms on the Cityscapes val set. 
    All the experiments are conducted with ResNet-101 as the backbone.
}
    \label{tab:cityscape}
\begin{tabular}{lcccc} \toprule 
Method & 1/16 (186) & 1/8 (372) & 1/4 (744) & 1/2 (1488) \\
\midrule Supervised & 65.7 & 72.5 & 74.4 & 77.8 \\ 
CCT~\cite{ouali2020semi} & 69.3 & 74.1 & 76.0 & 78.1 \\
CPS~\cite{chen2021semi} & 69.8 & 74.3& 74.6 & 76.8 \\
AEL~\cite{hu2021semi} & 74.5 & 75.5 & 77.5 & 79.0 \\
U$^2$PL~\cite{wang2022semi} & 70.3 & 74.4 & 76.5 & 79.1 \\
PS-MT~\cite{liu2022perturbed} & - & 76.9 & 77.6 & 79.1 \\
UniMatch\cite{yang2022revisiting} & 76.6 &77.9 & 79.2 & 79.5 \\
PCR~\cite{xu2022semi} & 73.4 & 76.3 & 78.4 & 79.1\\
\rowcolor[HTML]{EFEFEF} \textbf{\MyMthd{} } & \textbf{77.3}  &\textbf{78.5} & \textbf{79.4} & \textbf{80.4} \\
\bottomrule 
\end{tabular}
\end{table}

\myPara{Impact of label propagation strategies.} In \tabref{tab:propagation}, we conduct the ablation study of our label propagation strategies. Our pixel propagation strategy, which constructs the correlation maps and spreads them into predictions as a new representation with the supervision of correlation loss $\mathcal{L}^c$, brings 1.4\%, 0.4\%, and 0.8\% improvements. Furthermore, equipped with our region propagation strategy, more detailed local shape information is mined and thus enhanced pseudo labels are obtained. This strategy further improves 0.6\%, 0.5\%, and 0.5\% on 92, 366, and 1464 splits, respectively.

\myPara{Where to extract features.} In the default setting, we choose to extract features from the backbone, which makes the proposed strategies more convenient to be transplanted to other segmentation networks.  
Actually, given a specific network structure, the position of feature extraction can be flexible. 
Here, we consider the impact of different feature extraction positions on performance. 
In \tabref{feature}, we demonstrate the performance of extracting features after different positions for the Deeplabv3+ decoder under different splits.
The results show that using the backbone features consistently outperforms other alternatives.

\begin{table}[t]
      \small
      \centering
      \setlength{\abovecaptionskip}{4pt}
      \setlength{\tabcolsep}{12pt}
    \caption{
    Ablation study on the effectiveness of different components, including threshold $\tau$ (Dyna. denotes our dynamic strategy), hard loss $\mathcal{L}_{u}^{h}$, soft loss $\mathcal{L}_{u}^{s}$, label propagation $\mathcal{P}$.
    }
    \begin{tabular}{ccccccc} \toprule
     $\tau$ &$\mathcal{L}_{u}^{h}$ & $\mathcal{L}_{u}^{s}$ & $\mathcal{P}$  & 92 & 1464  \\
    \midrule
    Dyna. & \cmark & & &  73.6 &80.0\\
    Dyna. &  & \cmark & &  73.1&79.6\\
    Dyna. & \cmark & \cmark & & 74.4&80.5\\
    Dyna. & \cmark & & \cmark &  74.6 & 80.6\\
    Dyna. & \cmark & \cmark & \cmark &  $\mathbf{76.4}$ &$\mathbf{81.8}$\\
    \midrule
    Fixed & \cmark & & &  73.1&79.9\\
    Fixed & \cmark & \cmark &  & 73.3 &79.9\\
    Fixed & \cmark & & \cmark  &74.3 & 80.1\\
    Fixed & \cmark & \cmark & \cmark   & 75.5 & 80.8\\
    \bottomrule
    \end{tabular}
    \label{tab:components}
\end{table}

 \begin{table}[t]
\small
\centering
      \setlength{\abovecaptionskip}{4pt}
      \caption{Ablation study on the label propagation strategies. 
  }
  \setlength{\tabcolsep}{8.5pt}

  \begin{tabular}{lcccccc} 
    \toprule
    Method & 92  & 366  & 1464 \\
    \midrule
    w/o Propagation& 74.4 & 78.5&  80.5\\ 
    w/ Pixel Propagation& 75.8 & 78.9& 81.3\\ 
    w/ Pixel \& Region Propagation & \textbf{76.4} & \textbf{79.4}& \textbf{81.8}\\
    \bottomrule
  \end{tabular}
  \label{tab:propagation}
\end{table}

\myPara{Different sampling strategies.} Since using all shapes within the correlation map to enhance pseudo labels would incur a substantial computational burden, it is imperative to sample a subset of shapes from it. Here we conduct experiments about sampling methods and quantities in \tabref{tab:different sampling}.
We conduct experiments on random sampling $\mathcal{R}$ and uniform sampling $\mathcal{U}$ methods, with 16, 32, 64, 128, and 256 sampling numbers on the 1464 split. The results show random sampling continuously outperforms uniform sampling. Among these, random sampling with 128 sample numbers yields the best performance, with marginal differences compared to the 256-sample strategy. Thus, we choose to randomly sample 128 shapes from the correlation map as a trade-off between computational efficiency and performance.

 \myPara{Different initial values for \MyMthd{}.}
Since our EMA-based threshold updating strategy needs an initial value for $\tau$, we discuss the impact of different initialization values for $\tau$ in \figref{fig:different init}. 
The conclusion is that our threshold strategy is insensitive to different initialization values.
Even with different threshold initialization values, all the thresholds tend to approach a similar value very quickly (around 1500 iterations) in the early stage of training (around 40000 iterations in total) under all experiment settings.

 \begin{table}[t]
\small
\centering
      \setlength{\abovecaptionskip}{4pt}
      \caption{Ablation study on feature extraction positions. We take features after each specific module of DeepLabV3+ to build correlation maps and adopt label propagation strategies.
  }
  \setlength{\tabcolsep}{8.5pt}
  \begin{tabular}{ccccc} 
    \toprule
    Position  & Backbone & ASPP &  Fusion & Classifier \\
    \midrule
	732  & \textbf{80.4 } & 79.5 & 79.1 & 79.5 \\ 
        1464 & \textbf{81.8} & 80.6 &80.1 & 80.8\\
    \bottomrule
  \end{tabular}
  \label{feature}
\end{table}

 \begin{table}[t]
\small
\centering
      \setlength{\abovecaptionskip}{2pt}
      \caption{Ablation study on the different sampling methods. $\mathcal{R}$ denotes random sampling; $\mathcal{U}$ denotes uniform sampling.
  }
  \setlength{\tabcolsep}{10pt}

  \begin{tabular}{cccccc} 
    \toprule
    Numbers & 16  & 32 & 64 &  128 & 256 \\
    \midrule
	$\mathcal{R}$ & 81.1 & 81.2& 81.4& \textbf{81.8}& 81.7\\ 
        $\mathcal{U}$ & 81.0& 81.1 & 81.2 &81.4& 81.0\\
    \bottomrule
  \end{tabular}
  \label{tab:different sampling}
\end{table}

\begin{figure}[t]
\begin{center}
	\begin{subfigure}[t]{0.155\textwidth}
		\includegraphics[width=1\textwidth, height=0.9\textwidth]{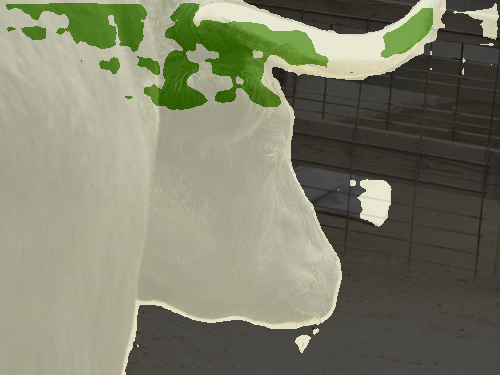}
	\end{subfigure}
	\begin{subfigure}[t]{0.155\textwidth}
		\includegraphics[width=1\textwidth, height=0.9\textwidth]{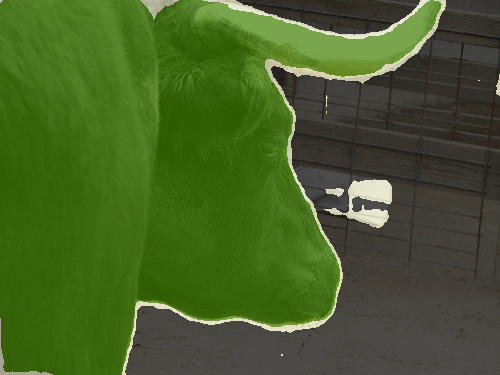}
	\end{subfigure}
 	\begin{subfigure}[t]{0.155\textwidth}
		\includegraphics[width=1\textwidth, height=0.9\textwidth]{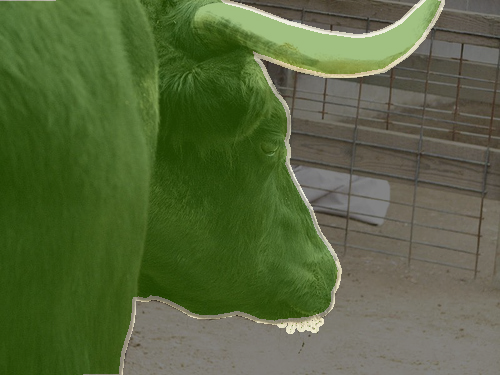}
	\end{subfigure}
 \\
	\begin{subfigure}[t]{0.155\textwidth}
		\includegraphics[width=1\textwidth, height=0.9\textwidth]{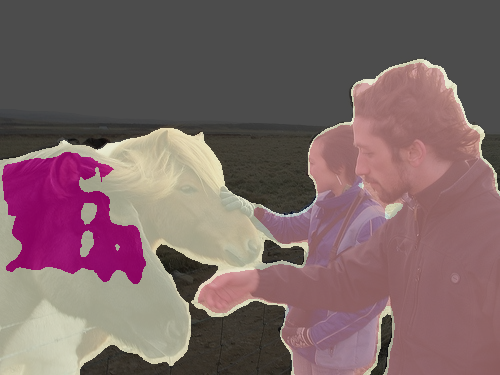}
	\end{subfigure}
	\begin{subfigure}[t]{0.155\textwidth}
		\includegraphics[width=1\textwidth, height=0.9\textwidth]{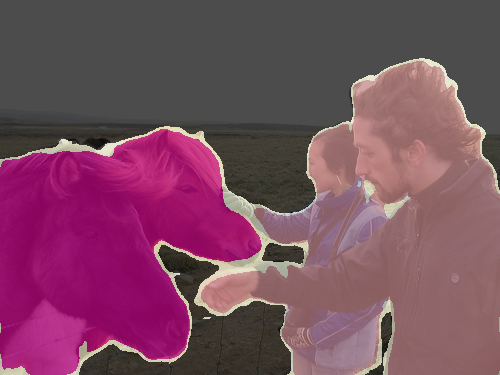}
	\end{subfigure}
 	\begin{subfigure}[t]{0.155\textwidth}
		\includegraphics[width=1\textwidth, height=0.9\textwidth]{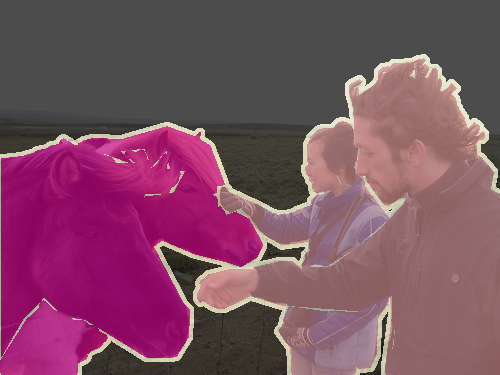}
	\end{subfigure}
 \\
	\begin{subfigure}[t]{0.155\textwidth}
		\includegraphics[width=1\textwidth, height=0.9\textwidth]{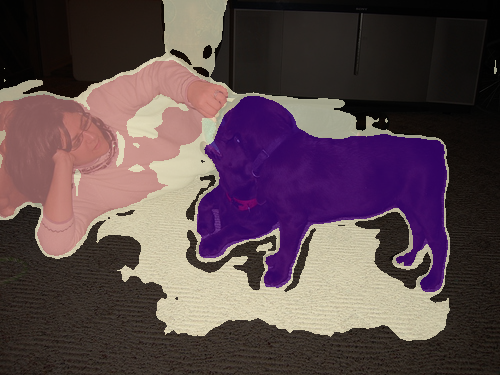}
        \subcaption{w/o propagation}
        \label{w/o corr}
	\end{subfigure}
	\begin{subfigure}[t]{0.155\textwidth}
		\includegraphics[width=1\textwidth, height=0.9\textwidth]{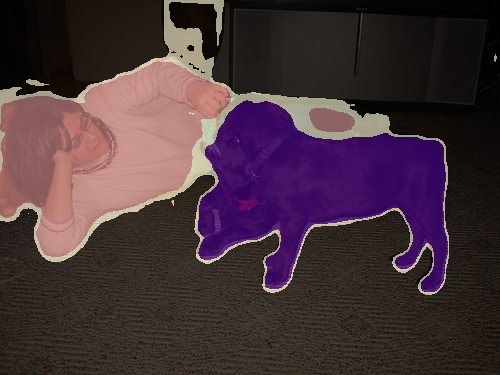}
        \subcaption{w/ propagation}
        \label{w/ corr}
	\end{subfigure}
 	\begin{subfigure}[t]{0.155\textwidth}
		\includegraphics[width=1\textwidth, height=0.9\textwidth]{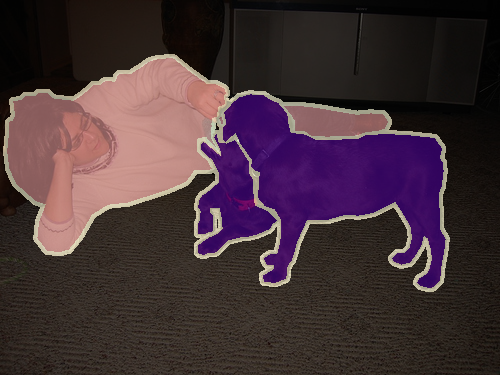}
        \subcaption{GT}
	\end{subfigure}
       \setlength{\abovecaptionskip}{2pt}
	\caption{Qualitative results on the Pascal VOC 2012 dataset. (a) Pseudo labels without label propagation; (b) Pseudo labels with \MyMthd{}; (c) Ground truth. White areas in (a) and (b) are ignored regions due to low confidence.}
	\label{fig:selected_pics}
\end{center}
\vspace{-15pt}
\end{figure}

\begin{figure*}
      \setlength{\abovecaptionskip}{2pt}
     \centering
         \begin{subfigure}[b]{0.24\textwidth}
         \centering
         \includegraphics[width=\textwidth]{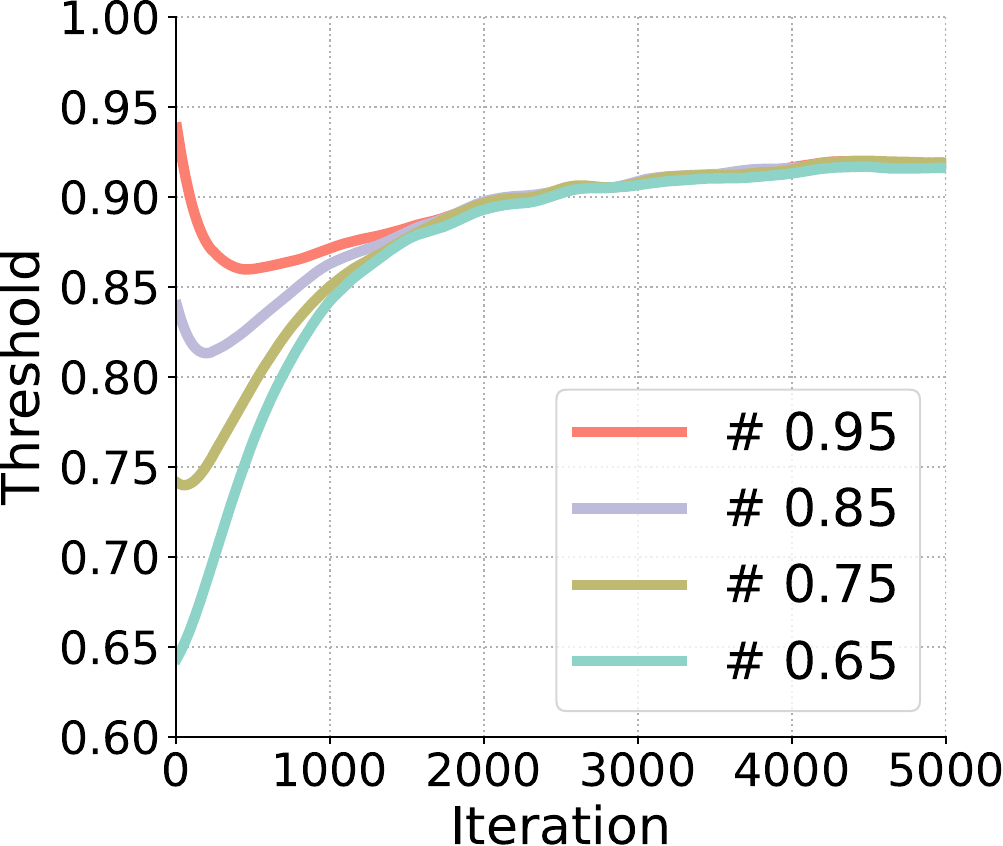}
         \caption{Different threshold initialization}
         \label{fig:different init}
     \end{subfigure}
     \hfill
    \begin{subfigure}[b]{0.24\textwidth}
         \centering
         \includegraphics[width=\textwidth]{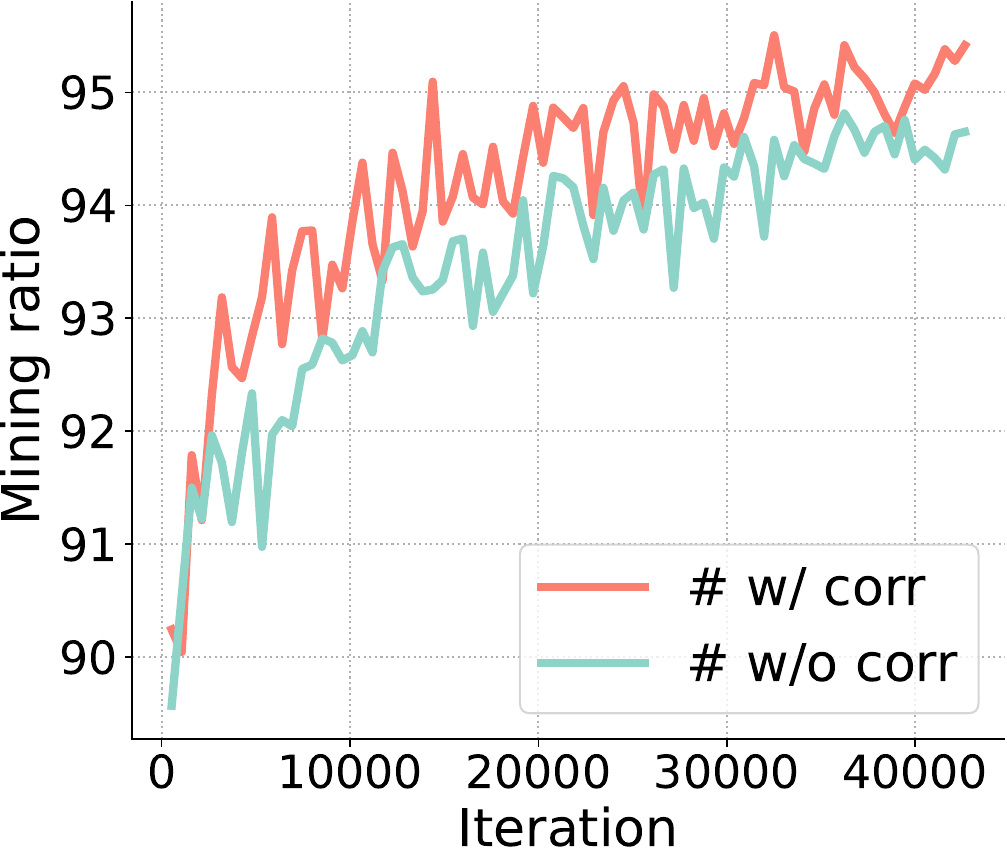}
         \caption{Mining ratio}
         \label{fig: mine ratio}
     \end{subfigure}
          \begin{subfigure}[b]{0.24\textwidth}
         \centering
         \includegraphics[width=\textwidth]{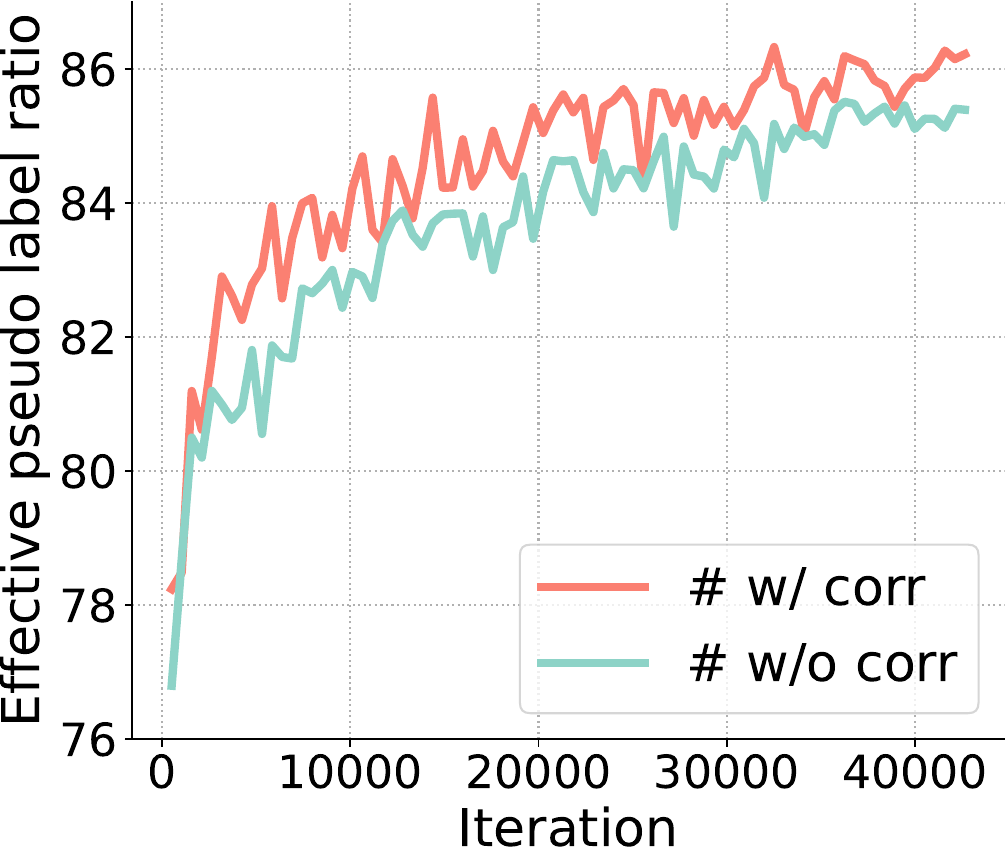}
         \caption{Effective pseudo label ratio}
         \label{fig: effective pseudo}
     \end{subfigure}
     \hfill
     \begin{subfigure}[b]{0.25\textwidth}
         \centering
         \includegraphics[width=\textwidth]{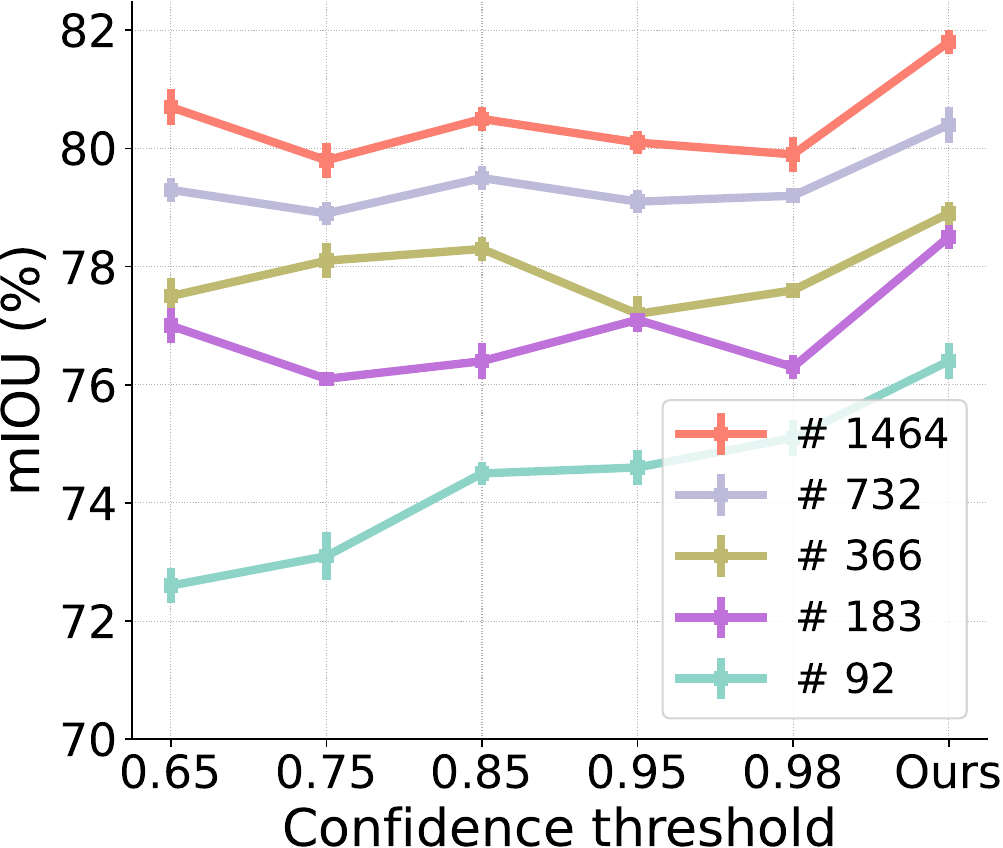}
         \caption{Different fixed thresholds}
         \label{fig:different fix}
     \end{subfigure}
     \hfill

        \caption{Some statistics on label propagation and the threshold strategy. For (a), (b), and (c), experiments are conducted on the 1464 split.}
        \label{fig:statistics}
    \vspace{-10pt}
\end{figure*}

\subsection{Correlation Helps Mining Reliable Regions}

\myPara{Statistics.} 
Ideally, all correctly predicted points should be regarded as pseudo labels for the unlabeled data. To demonstrate the ability of correlation matching to help label propagation, we count the mining ratio and effective pseudo label ratio in \figref{fig: mine ratio} and \figref{fig: effective pseudo}. The mining ratio is the proportion of selected high-confidence pixels among all correctly predicted pixels. The effective pseudo label ratio is the proportion of accurately predicted pseudo labels to the whole image, which can reflect effective pseudo label numbers. 
It can be clearly seen that with the proposed label propagation strategies, the mining ratio and effective pseudo label ratio are significantly higher than those without them, which illustrates that the utilization of unlabeled data has improved effectively.
This further indicates our strategies can improve the overall quality of pseudo labels by leveraging similarity and shape information from correlation maps.

\myPara{Qualitative analysis.}
In \figref{fig:selected_pics}, we give some visualization results to further demonstrate the effectiveness of our label propagation strategies. 
Comparing \figref{w/ corr} and \figref{w/o corr}, it is obvious that with the support of label propagation, the number of pixels and completeness of the high-confidence regions are significantly better than those without it. This means that our method can effectively expand high-confidence regions and populate these regions with the correct categories.
%
%
%
%
%
We will provide more detailed qualitative results in the supplementary materials.

\section{Discussion about Label propagation Strategy v.s. Threshold Adjustment }
\begin{figure}
\begin{center}
         \begin{subfigure}[t]{0.235\textwidth}
         \includegraphics[width=\textwidth]{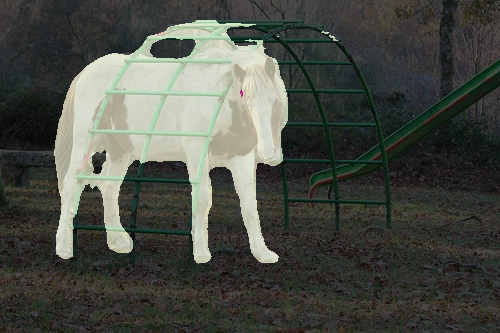}
         \caption{Threshold=0.95}
         \label{fig:high}
     \end{subfigure}
    \begin{subfigure}[t]{0.235\textwidth}
         \includegraphics[width=\textwidth]{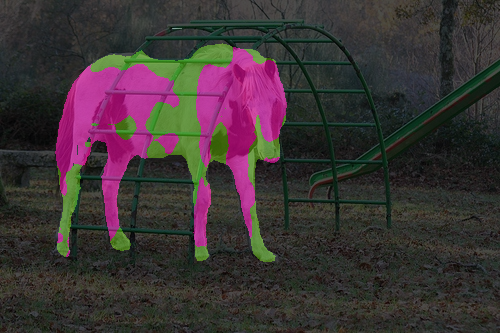}
         \caption{Threshold=0}
         \label{fig:low}
     \end{subfigure}
\\
    \begin{subfigure}[t]{0.235\textwidth}
         \includegraphics[width=\textwidth]{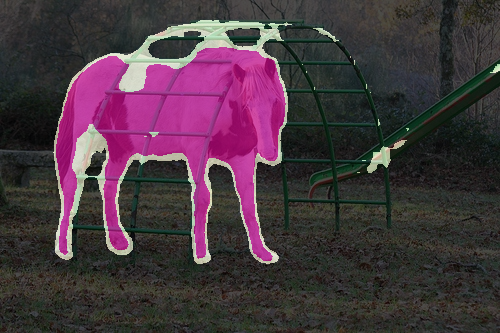}
         \caption{Label propagation}
         \label{fig:propagation}
     \end{subfigure}
     \begin{subfigure}[t]{0.235\textwidth}
         \includegraphics[width=\textwidth]{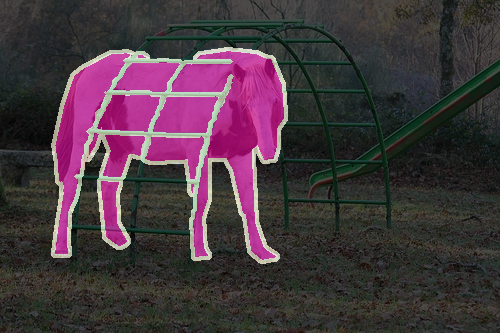}
         \caption{GT}
         \label{fig:gt}
     \end{subfigure}
    \setlength{\abovecaptionskip}{2pt}
        \caption{Comparisons of pseudo labels with different strategies.}
        \label{fig:pseudo labels}
        \vspace{-20pt}
\end{center}
\end{figure}

Traditionally, semi-supervised semantic segmentation methods mostly rely on adjusting thresholds to expand high-confidence regions~\cite{yang2022revisiting, wang2022semi}. However, selecting the most suitable threshold could be a challenging task. For instance, our observations illustrated in \figref{fig:different fix}, indicate that the optimal threshold can vary significantly.  \figref{fig:high} and \figref{fig:low} further demonstrate that a too-strict threshold restricts the unlabeled data utilization, while a lenient threshold results in fragmented incorrect pixel predictions. 

Different from the scheme of directly adjusting the threshold, label propagation does not merely expand the high-confidence regions; it assigns accurate predictions to pseudo labels by utilizing accurate shapes within the correlation map, which helps maintain more consistent semantic structures within high-confidence regions and thus mitigates the discontinuity issue.
%
In  \figref{fig:propagation} and the last column of \figref{fig:different fix}, we show the pseudo label and performance of \MyMthd{}.
%
%
It indicates that our \MyMthd{} consistently obtains more accurate and complete pseudo labels and achieves the highest results on all splits, demonstrating the effectiveness of the proposed label propagation strategies.

\section{Conclusions}
We present \MyMthd{} that can utilize label propagation with correlation matching to discover more accurate high-confidence regions for semi-supervised semantic segmentation. 
The key contributions of our \MyMthd{} are reconsidering the use of correlation maps and designing two label propagation strategies to enrich the pseudo label. Equipped with these strategies, CorrMatch significantly expands the high-confidence regions and thus can utilize unlabeled data more efficiently.
%
Experiments show the superiority of our \MyMthd{} over other methods.


\newpage
\appendix
\section{Pseudocode for proposed strategies}
\subsection{Pseudocode for Region Propagation}
In \secref{sec: region prop} of our main paper, we propose the region propagation strategy. This strategy combines the
shape information sampled from correlation maps with the most salient class to enhance the pseudo label and expand the high-confidence regions.
Here we present the pseudocode of the region propagation strategy in a PyTorch-like style.

\begin{algorithm}[H]
\caption{Pseudocode of region propagation strategy in a PyTorch-like style.}
\label{alg:region propagation}
\definecolor{codeblue}{rgb}{0.25,0.5,0.5}
\lstset{
  backgroundcolor=\color{white},
  basicstyle=\fontsize{8pt}{8pt}\ttfamily\selectfont,
  columns=fullflexible,
  breaklines=true,
  captionpos=b,
  commentstyle=\fontsize{8pt}{8pt}\color{Green},
  keywordstyle=\fontsize{8pt}{8pt}\bfseries,
  emph={unique, argmax, :}, emphstyle=\bfseries,
}
\begin{lstlisting}[language=python]
# shapes: Binary shape information sampled from correlation maps
# t: Confidence threshold
# hc_regions: Current high-confidence regions
# pseudo_label: Current pseudo label
def Region(shapes, t, hc_regions, pseudo_label):
    # Find the high-confidence shapes
    hc_shapes = shapes * hc_regions 
    b, c, h, w = shapes.shape
    
    for i in range(b):
        for j in range(c):
            hc_shape = hc_shapes[i, j]
            shape = shapes[i, j]
            
            #  Calculate the overlap between the high-confidence shape and original shape
            r1 = sum(hc_shape) / sum(shape)
            if r1 < t:
                continue    
                
            # Find all unique classes and their counts in the pseudo label within the high-confidence shape
            unique_cls, cnt = unique(pseudo_label[i][hc_shape == 1])
            
            # Calculate the ratio of the most salient class within the high-confidence shape
            r2 = max(cnt) / sum(cnt)
            if r2 < t:
                continue
            
            # Assign the most salient class to the pseudo label with shape information
            top_cls = unique_cls[argmax(cnt)]
            pseudo_label[i][shape == 1] = top_cls
    
            # Update the new high-confidence regions with the current shape
            hc_regions[i] = hc_regions[i] | shape
\end{lstlisting}
\end{algorithm}

\subsection{Pseudocode for Threshold Updating}
In \secref{sec: more details} of our main paper, we propose the threshold updating strategy. Our core idea is maintaining a dynamic global threshold related to the model's learning process. Specifically, during the optimization process, we gradually update the threshold using the average of the maximum confidence of all predicted classes in weakly augmented predictions. With the increment $\Delta\tau$ proposed in Eqn (11) of our main paper, the EMA procedure is defined as:
\begin{equation}
\label{eqn: ema}
\tau= \lambda \tau + (1-\lambda) \Delta\tau,
\end{equation}
where $\lambda$ is the momentum decay of EMA.
To make things more clear, we here present the pseudocode of the threshold updating strategy in a PyTorch-like style.

\begin{algorithm}[H]
\caption{Pseudocode of threshold updating strategy in a PyTorch-like style.}
\label{alg:relaxmatch}
\definecolor{codeblue}{rgb}{0.25,0.5,0.5}
\lstset{
  backgroundcolor=\color{white},
  basicstyle=\fontsize{8pt}{8pt}\tt\selectfont,
  columns=fullflexible,
  breaklines=true,
  captionpos=b,
  commentstyle=\fontsize{8pt}{8pt}\color{ForestGreen},
  keywordstyle=\fontsize{8pt}{8pt}\bfseries,
  emph={unique, argmax, softmax, :}, emphstyle=\bfseries,
}
\begin{lstlisting}[language=python]
# pred: Logits of weak augmented images 
# thresh_global: Current global threshold
# momentum: Coefficient of EMA
def update(pred, thresh_global, momentum):
    # initialize update value
    update_value = 0.0
    
    # get predicted mask and confidence from pred
    mask_pred = argmax(pred, dim=1)
    pred_conf = pred.softmax(dim=1).max(dim=1)
    
    # find all classes in the predicted mask
    unique_cls = unique(mask_pred)
    cls_num = len(unique_cls)
    
    for cls in unique_cls:
        # find the highest confidence score for each predicted class
        cls_map = (mask_pred == cls)
        pred_conf_cls_all = pred_conf[cls_map]
        cls_max_conf = pred_conf_cls_all.max()
        update_value += cls_max_conf
    
    # get the mean of all confidence scores
    update_value = update_value / cls_num
    
    # update thresh_global in EMA style
    thresh_global = momentum * thresh_global + (1 - momentum) * update_value
\end{lstlisting}
\end{algorithm}

\section{More Implementation Details}
\myPara{Data augmentations.} We followed the common settings from previous works~\cite{yang2022st++, yang2022revisiting, zou2020pseudoseg}. For weak data augmentation, we use the random scale with a range [0.5, 2.0], the random horizontal flip with a probability of 0.5, and the random crop with a certain size (321, 513, or 801). As for strong data augmentation, we use the colorjitter technique to change the brightness, contrast, saturation, and hue of the image with the same parameter setting as previous works~\cite{yang2022st++, yang2022revisiting, zou2020pseudoseg}. Random grayscale and gaussian blur are also applied as strong data augmentations.
We also use the CutMix~\cite{yun2019cutmix} technique as done in many previous approaches~\cite{yang2022st++, yang2022revisiting, zou2020pseudoseg}. 
Besides, to learn more robust feature representations, we use the same feature perturbations (randomly dropout 50\% of the channels from the encoder feature) as UniMatch~\cite{yang2022revisiting}.

\myPara{Feature extractor.} As mentioned in \secref{sec:correlation-maps} of our main paper, we extract features from the encoder of the network. The specific extractor comprises a $3\times3$ convolution, followed by batch normalization~\cite{ioffe2015batch} and an activation layer. Then, two individual linear transformations are adopted on the extracted feature to obtain the $w_1$ and $w_2$.

\myPara{Others.} We use the stochastic gradient descent (SGD) optimizer with momentum $=0.9$ and the poly scheduling with $(1 - \frac{\text{iter}}{\text{total iter}})^{0.9}$ to decay the learning rate during the training process. Furthermore, we set the momentum of EMA to 0.999 for the proposed dynamic threshold updating strategy. And same to UniMatch~\cite{yang2022revisiting}, we set the confidence threshold $\tau$ to 0 for the Cityscapes dataset.

\section{More Ablation Studies}
\subsection{Impact of momentum decay}
Considering that \MyMthd{} uses EMA to iteratively update dynamic thresholds, in \tabref{tab: ema decay}, we perform ablation experiments on the momentum decay of EMA.
\begin{table}[t]
    \centering
        \small
    \setlength{\tabcolsep}{16pt}
    \caption{ Comparison of \MyMthd{} with different momentum decay of EMA on PASCAL VOC 2012 val set with mIoU (\%) $\uparrow$ metric.  }
    \label{tab: ema decay}
\begin{tabular}{ccc} \toprule
 momentum decay & 1 / 16(92)  & Full (1464) \\
\midrule 
0.99 & 75.6  & 79.8\\
0.999& \textbf{76.4}  & \textbf{81.8} \\
0.999 & 75.7  & 80.3\\
\bottomrule 
\end{tabular}
\end{table}
\subsection{Different Soft Supervision}
As mentioned in \secref{framework}, we introduce soft supervision into semi-supervised semantic segmentation. In \tabref{tab: different soft} we conduct experiments involving some different soft supervision techniques, and their similar results indicate that KL divergence is just a soft measurement and alternative soft supervision can achieve comparable performance.

\subsection{Different loss weights}
In \tabref{tab: different loss weights}, We conduct more ablation experiments on different loss weights. When the weight assigned to unlabeled data is excessively large, it significantly affects the model's performance, whereas more balanced weights have a minor impact on the model's performance. The results show that setting [$\lambda_1$, $\lambda_2$, $\lambda_3$] to [0.5, 0.25, 0.25] achieves the best performance.

\section{More Analysis for Label Propagation}

\subsection{Correlation module is not an attention module.}
The construction of correlation maps differs from the attention mechanism, exhibiting fundamental distinctions.
\begin{itemize}
    \item Formally, in the attention mechanism, both the key (K) and value (V) are derived from the same feature representations, often within the same input sequence. In contrast, our correlation mechanism first calculates the correlation map between the extracted feature representations and then the pixel propagation strategy is adopted to spread them into model output, which is obviously different sources from the extracted features.
    \item As for correlation maps and attention maps, correlation maps encode pairwise similarity between features from different regions, while attention maps are a set of weights that determine the importance of different positions in the input sequence.
\end{itemize}
In summary, our correlation module differs from the attention mechanism in terms of both form and encoded content. Besides, the proposed two label propagation strategies involve propagating the correlation maps to the output and enhancing the pseudo label with shape information, making our correlation module different from the attention module.

\begin{table}[t]
    \centering
        \small
    \setlength{\tabcolsep}{9pt}
    \caption{ Comparison of \MyMthd{} with different soft supervision on PASCAL VOC 2012 val set with mIoU (\%) $\uparrow$ metric.}
    \label{tab: different soft}
\begin{tabular}{lcc} \toprule
 Method & 1 / 16(92)  & Full (1464) \\
\midrule 
Kullback-Leibler divergence & 76.4  & 81.8\\
Soft cross-entropy& 76.2  & 81.6 \\
Cosine similarity & 76.1  & 81.5\\
\bottomrule 
\end{tabular}
\end{table}

\begin{table}[t]
    \centering
        \small
    \setlength{\tabcolsep}{17pt}
    \caption{ Comparison of \MyMthd{} with different loss weights on PASCAL VOC 2012 val set with mIoU (\%) $\uparrow$ metric.}
    \label{tab: different loss weights}
\begin{tabular}{lcc} \toprule
  $ [\lambda_1 , \lambda_2 , \lambda_3 ]$& 1 / 16(92)  & Full (1464) \\
\midrule 
$[0.5, 0.25, 0.25]$& \textbf{76.4}  & \textbf{81.8}\\
$[0.25, 0.5, 0.25]$& 75.6  & 81.2 \\
$[0.25, 0.25, 0.5]$ & 75.9  & 81.1\\
$[0.3, 0.3, 0.3]$ & 75.4  & 80.2\\
$[0.5, 0.5, 0.5]$ & 73.4  & 79.5\\
$[1, 1, 1 ]$& 70.6  & 78.0\\
\bottomrule 
\end{tabular}
\end{table}


\subsection{More Statistics}

\begin{figure}[t]
     \centering
    \begin{subfigure}[b]{0.23\textwidth}
         \centering
         \includegraphics[width=\textwidth]{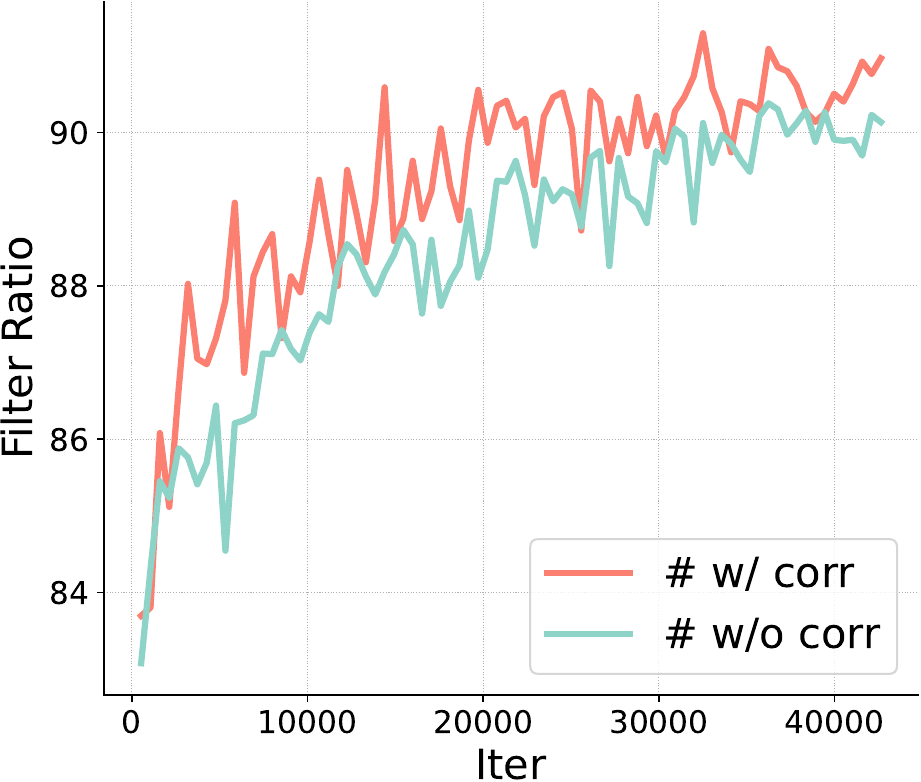}
         \caption{Filter Ratio}
         \label{fig: filter ratio}
     \end{subfigure}
     \hfill
\hfill
    \begin{subfigure}[b]{0.23\textwidth}
         \centering
         \includegraphics[width=\textwidth]{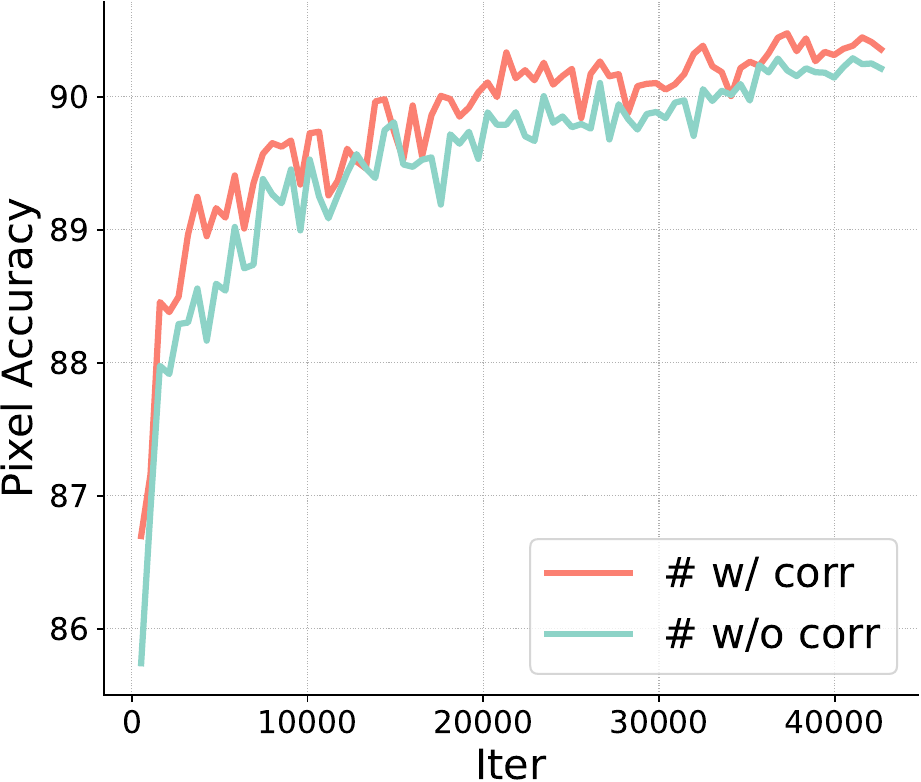}
         \caption{Pixel Accuracy}
         \label{fig: pixel accuarcy}
     \end{subfigure}
        \caption{More statistics about label propagation strategies.}
        \label{fig: more statistics}
\end{figure}

In \figref{fig: more statistics}, we demonstrate more statistics on the val set of the Pascal VOC 2012 dataset to further show the effectiveness of label propagation via correlation matching. 
We further count the filter ratio, and pixel accuracy with and without adopting correlation matching in \figref{fig: filter ratio} and \figref{fig: pixel accuarcy}, respectively.
The filter ratio is the proportion of high-confidence pixels that are regarded as pseudo-labels for the whole image, which can reflect the overall confidence of the model.
And the pixel accuracy is all accurately predicted pixels to the whole image.  All the experiments are conducted on the 1464 split with training size 321$\times$321.

It can be clearly seen that the trend of the two curves in these three figures is consistent.
That is, using correlation matching can yield much better results. This means that not only does the model tend to make predictions with overall higher confidence, but the number of high-confidence pixels that are correctly predicted increases.
Also, higher pixel accuracy with correlation matching indicates better performance of the model itself.
These statistics further demonstrate that our proposed \MyMthd{} with the label propagation strategy can mine more accurate high-confidence regions and thus boost the model to learn more from the unlabeled data.

\begin{figure*}
     \centering
     \begin{subfigure}[b]{0.23\textwidth}
         \centering
         \includegraphics[width=\textwidth]{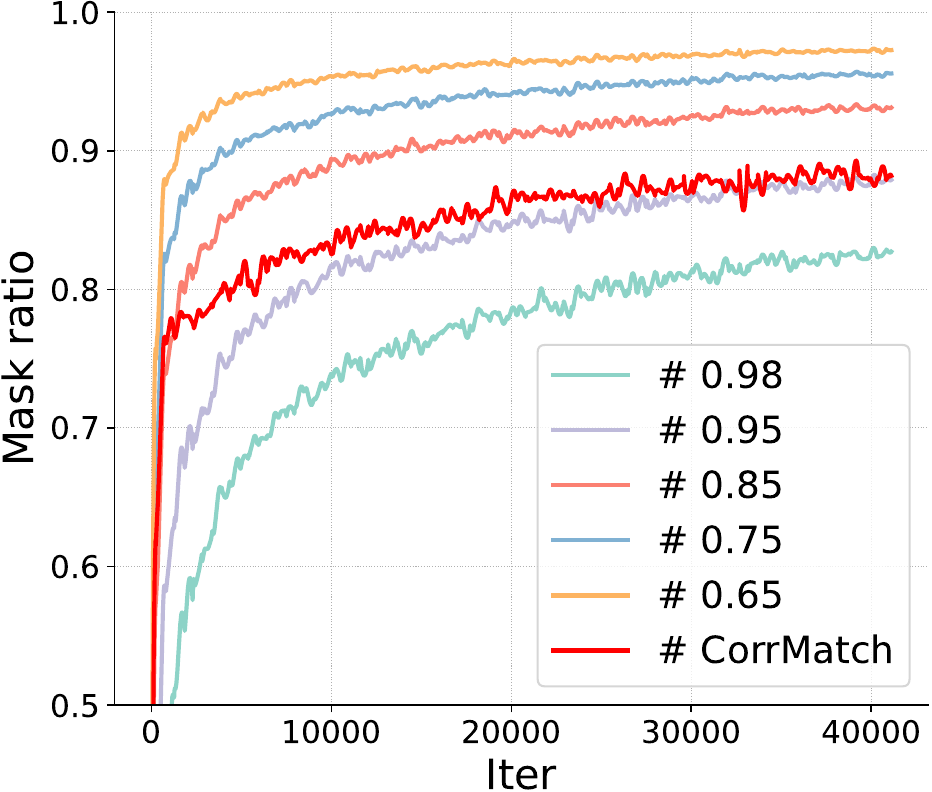}
         \caption{1464 split}
         \label{fig: mask ratio 1464}
     \end{subfigure}
     \hfill
     \begin{subfigure}[b]{0.23\textwidth}
         \centering
         \includegraphics[width=\textwidth]{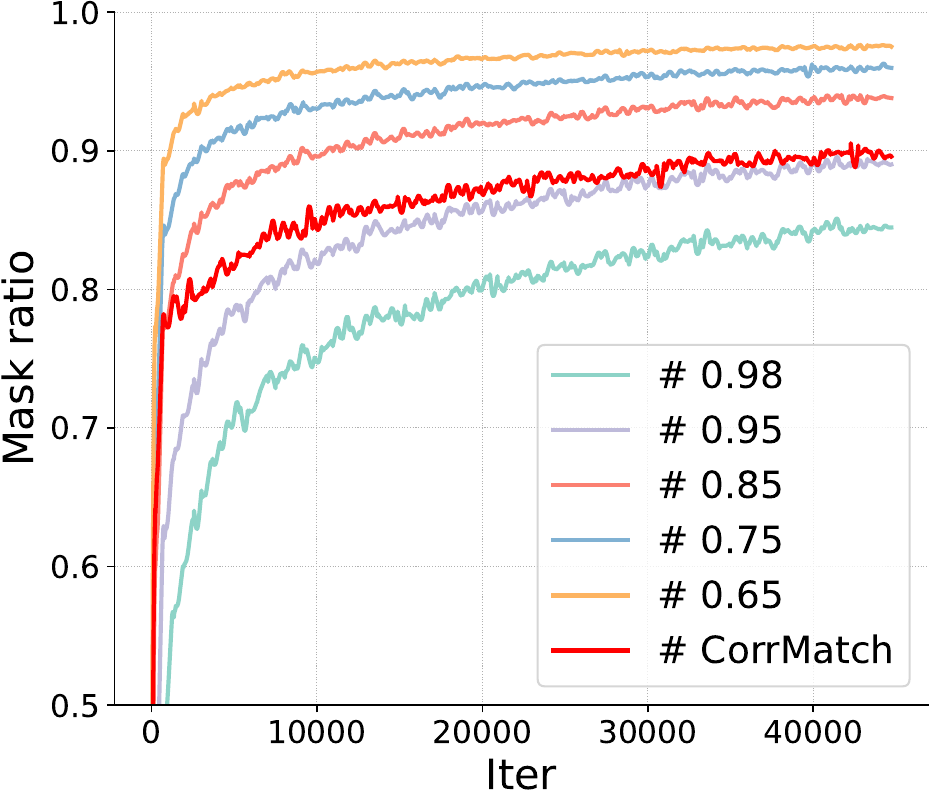}
         \caption{732}
         \label{fig: mask ratio 732}
     \end{subfigure}
     \hfill
     \begin{subfigure}[b]{0.23\textwidth}
         \centering
         \includegraphics[width=\textwidth]{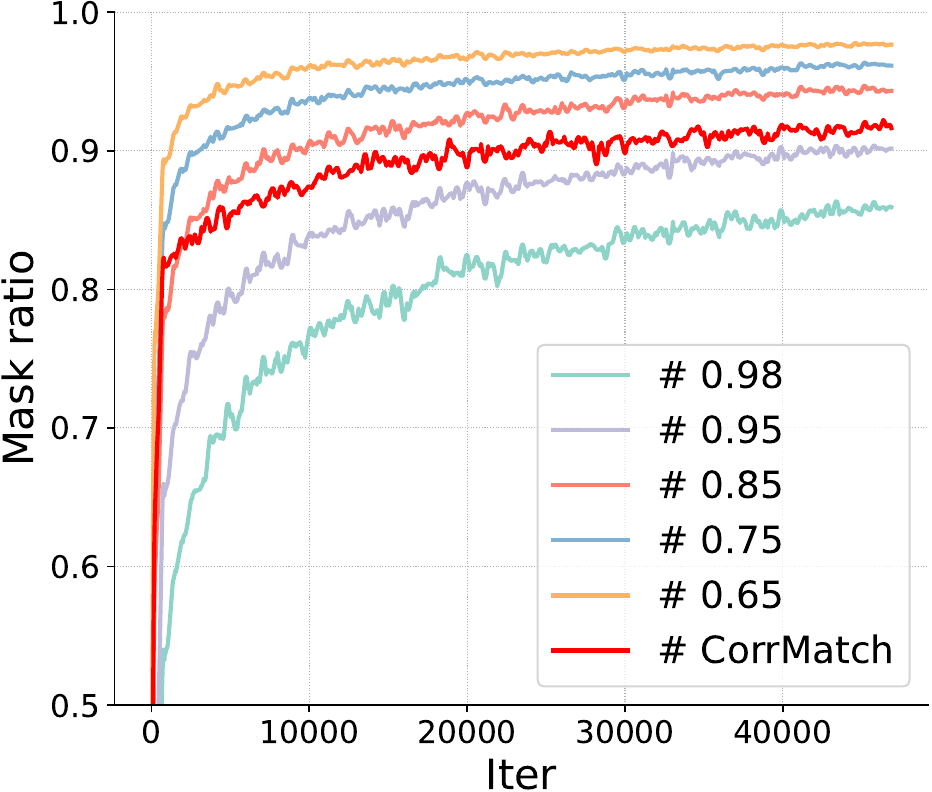}
         \caption{366 split}
         \label{fig: mask ratio 366}
     \end{subfigure}
     \hfill
     \begin{subfigure}[b]{0.23\textwidth}
         \centering
         \includegraphics[width=\textwidth]{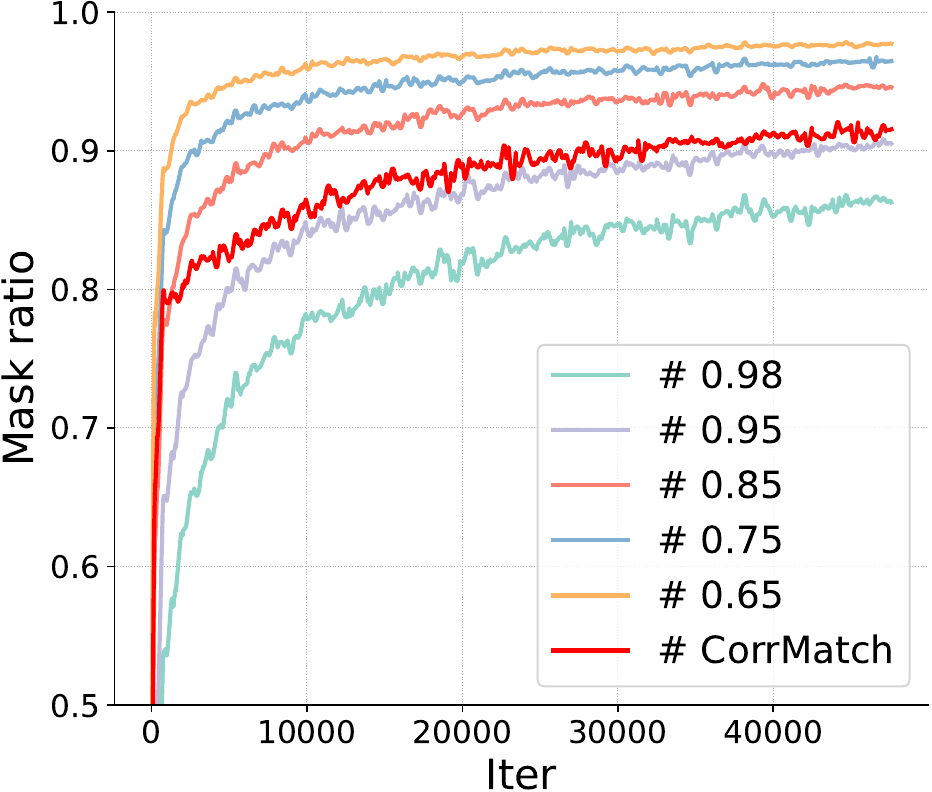}
         \caption{92 split}
         \label{fig: mask ratio 92}
     \end{subfigure}
        \caption{Mask ratio during the training process of different splits with different fixed thresholds.}
        \label{fig: mask ratio}
\end{figure*}

\subsection{Mask Ratio}
In \figref{fig: mask ratio}, we demonstrate the mask ratio (proportion of high-confidence pixels filtered by the threshold) during the training process. We compare the mask ratio statistics using fixed thresholds with using our \MyMthd{}. 
It is obvious that the lower the fixed threshold is, the higher the mask ratio will be. Moreover, a too-low mask ratio in the early training will lead to fewer predictions that constitute pseudo-labels, which will affect the convergence speed.
On the contrary, a too-high mask ratio in the later training will contain more wrong predictions, which will affect the accuracy of pseudo-labels.
Both situations are detrimental to model convergence.
However, our \MyMthd{} tackles this problem by achieving a relatively higher mask ratio early and a relatively lower mask ratio later. This phenomenon maintains a consistent trend in \figref{fig: mask ratio 1464}, \figref{fig: mask ratio 366}, \figref{fig: mask ratio 732}, and \figref{fig: mask ratio 92}, thus further verifying the stability of our method.

\subsection{More Visualizations}
\begin{figure*}[ht]
\begin{center}
  \begin{subfigure}[t]{0.135\textwidth}
		\includegraphics[width=1\textwidth, height=0.9\textwidth]{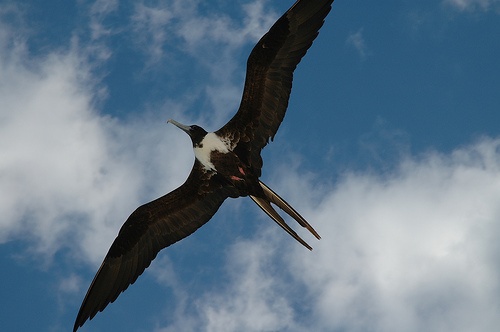}
	\end{subfigure}
	\begin{subfigure}[t]{0.135\textwidth}
		\includegraphics[width=1\textwidth, height=0.9\textwidth]{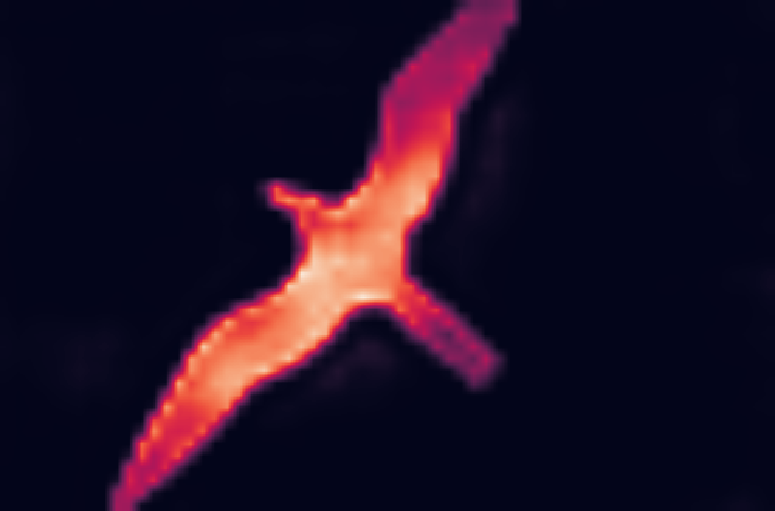}
	\end{subfigure}
	\begin{subfigure}[t]{0.135\textwidth}
		\includegraphics[width=1\textwidth, height=0.9\textwidth]{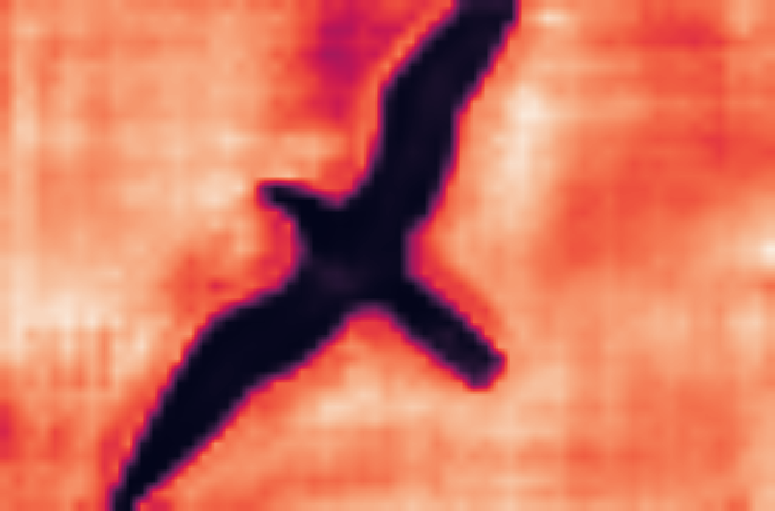}
	\end{subfigure}
	\begin{subfigure}[t]{0.135\textwidth}
		\includegraphics[width=1\textwidth, height=0.9\textwidth]{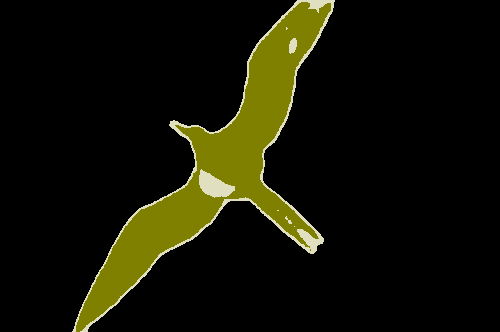}
	\end{subfigure}
	\begin{subfigure}[t]{0.135\textwidth}
		\includegraphics[width=1\textwidth, height=0.9\textwidth]{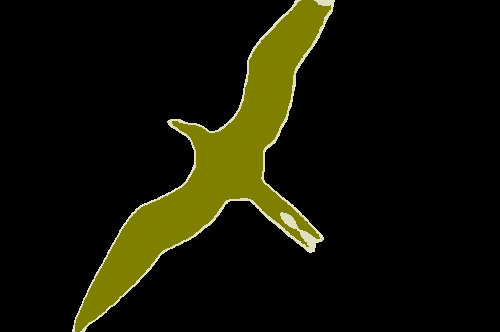}
	\end{subfigure}
 	\begin{subfigure}[t]{0.135\textwidth}
		\includegraphics[width=1\textwidth, height=0.9\textwidth]{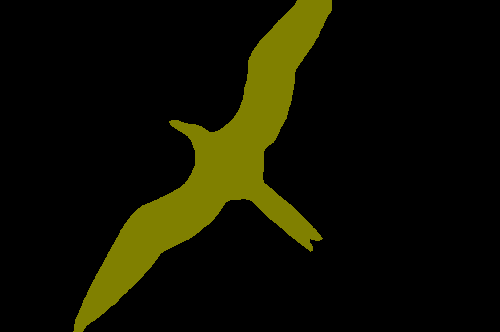}
	\end{subfigure}
 	\begin{subfigure}[t]{0.135\textwidth}
		\includegraphics[width=1\textwidth, height=0.9\textwidth]{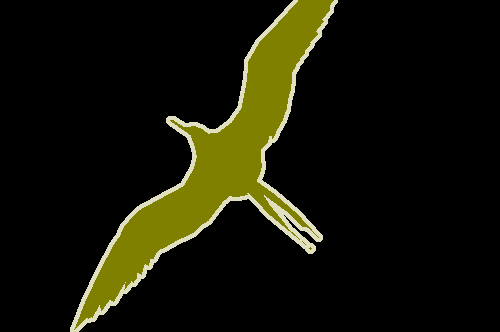}
	\end{subfigure}

  \begin{subfigure}[t]{0.135\textwidth}
		\includegraphics[width=1\textwidth, height=0.9\textwidth]{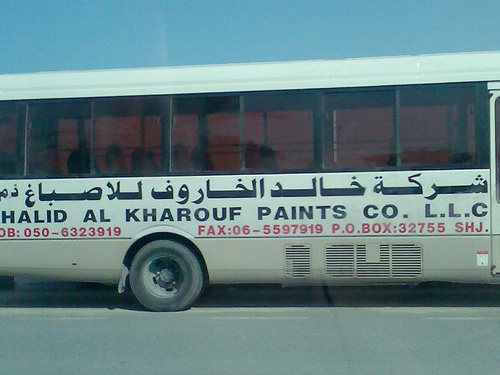}
	\end{subfigure}
	\begin{subfigure}[t]{0.135\textwidth}
		\includegraphics[width=1\textwidth, height=0.9\textwidth]{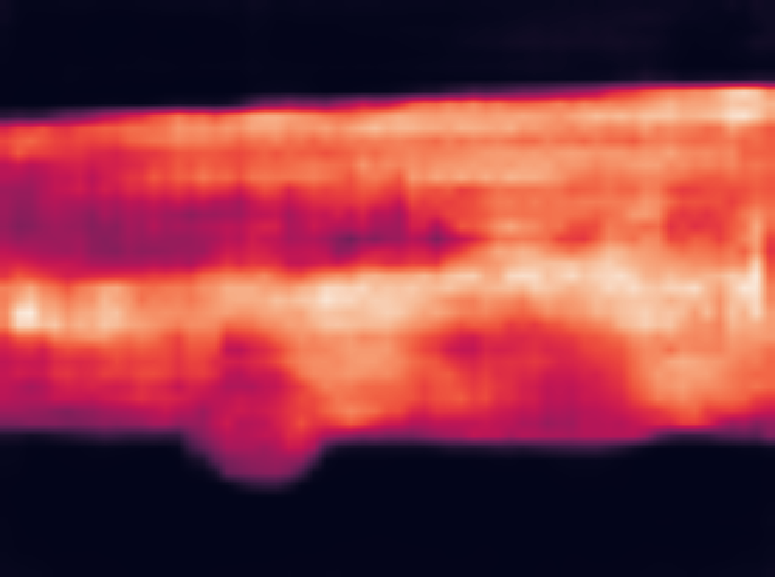}
	\end{subfigure}
	\begin{subfigure}[t]{0.135\textwidth}
		\includegraphics[width=1\textwidth, height=0.9\textwidth]{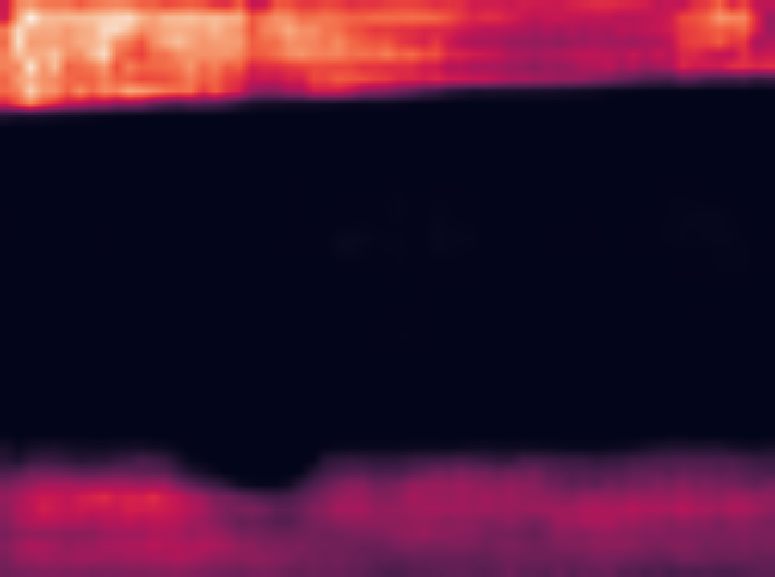}
	\end{subfigure}
	\begin{subfigure}[t]{0.135\textwidth}
		\includegraphics[width=1\textwidth, height=0.9\textwidth]{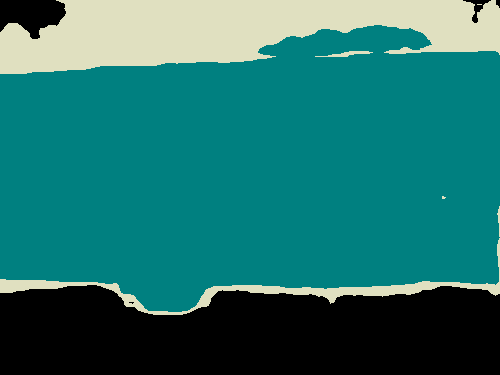}
	\end{subfigure}
	\begin{subfigure}[t]{0.135\textwidth}
		\includegraphics[width=1\textwidth, height=0.9\textwidth]{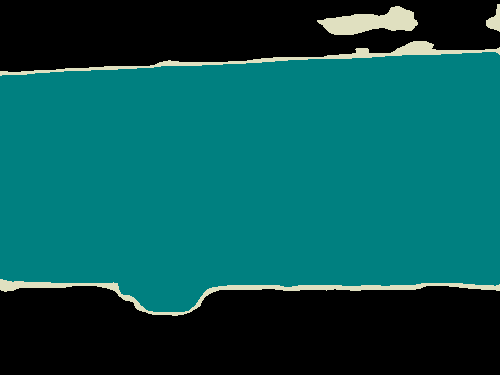}
	\end{subfigure}
 	\begin{subfigure}[t]{0.135\textwidth}
		\includegraphics[width=1\textwidth, height=0.9\textwidth]{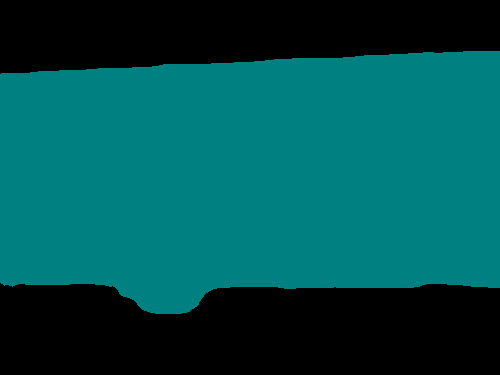}
	\end{subfigure}
 	\begin{subfigure}[t]{0.135\textwidth}
		\includegraphics[width=1\textwidth, height=0.9\textwidth]{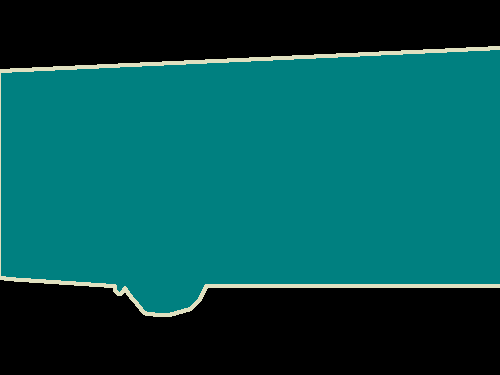}
	\end{subfigure}

    \begin{subfigure}[t]{0.135\textwidth}
		\includegraphics[width=1\textwidth, height=0.9\textwidth]{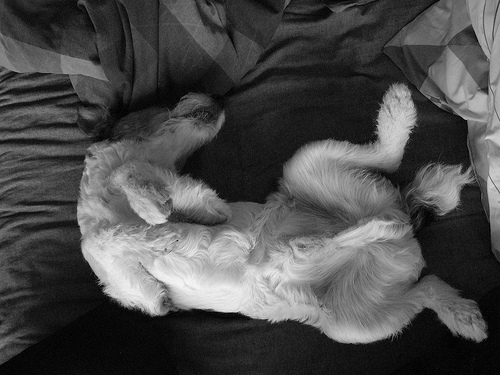}
	\end{subfigure}
	\begin{subfigure}[t]{0.135\textwidth}
		\includegraphics[width=1\textwidth, height=0.9\textwidth]{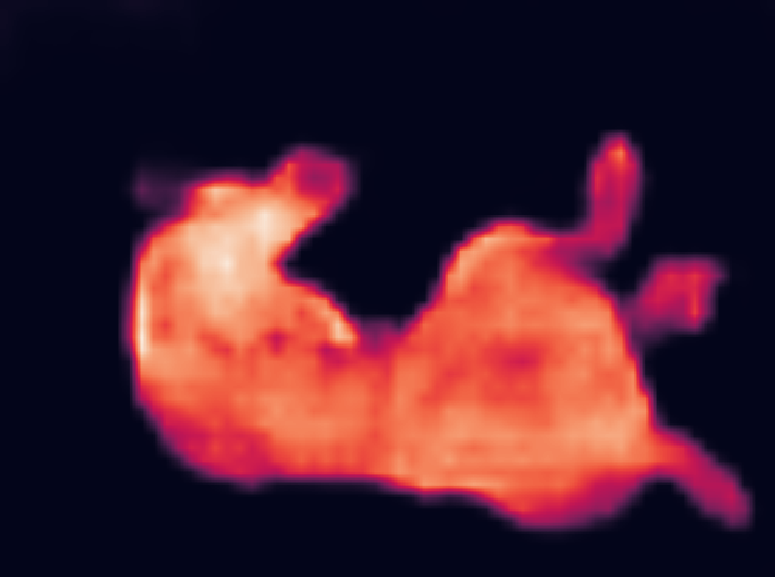}
	\end{subfigure}
	\begin{subfigure}[t]{0.135\textwidth}
		\includegraphics[width=1\textwidth, height=0.9\textwidth]{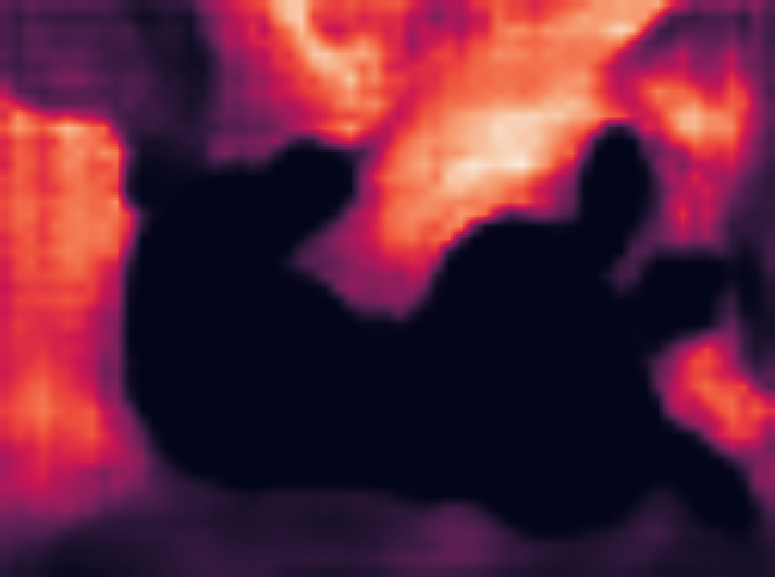}
	\end{subfigure}
	\begin{subfigure}[t]{0.135\textwidth}
		\includegraphics[width=1\textwidth, height=0.9\textwidth]{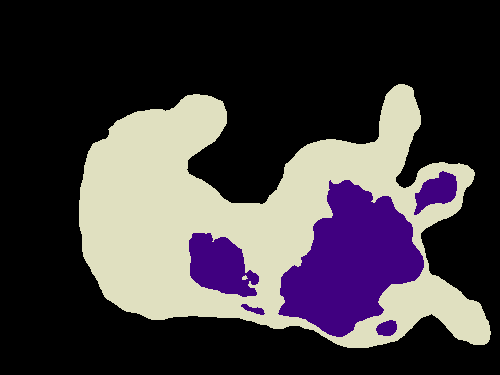}
	\end{subfigure}
	\begin{subfigure}[t]{0.135\textwidth}
		\includegraphics[width=1\textwidth, height=0.9\textwidth]{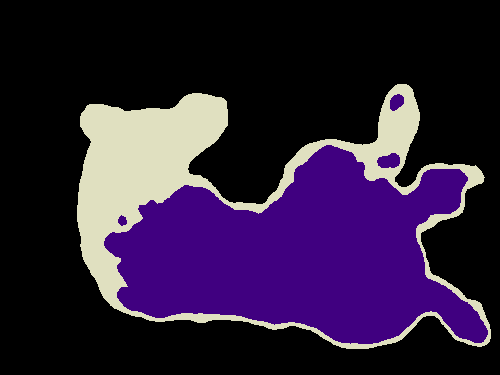}
	\end{subfigure}
 	\begin{subfigure}[t]{0.135\textwidth}
		\includegraphics[width=1\textwidth, height=0.9\textwidth]{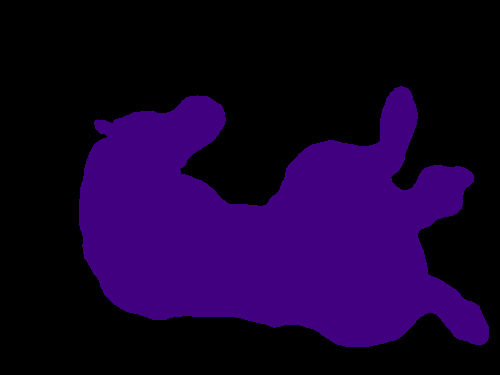}
	\end{subfigure}
 	\begin{subfigure}[t]{0.135\textwidth}
		\includegraphics[width=1\textwidth, height=0.9\textwidth]{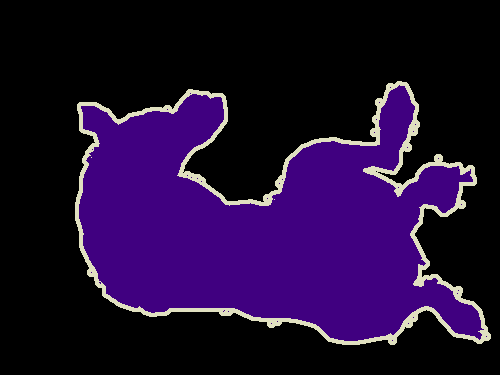}
	\end{subfigure}
  \begin{subfigure}[t]{0.135\textwidth}
		\includegraphics[width=1\textwidth, height=0.9\textwidth]{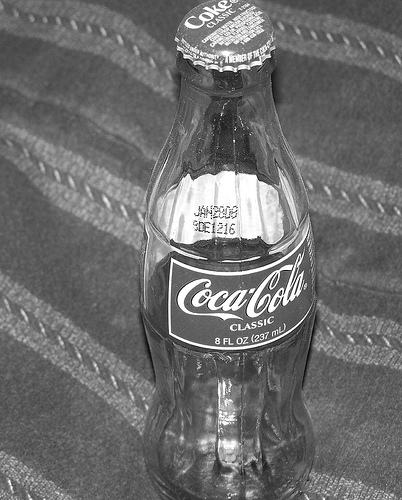}
	\end{subfigure}
	\begin{subfigure}[t]{0.135\textwidth}
		\includegraphics[width=1\textwidth, height=0.9\textwidth]{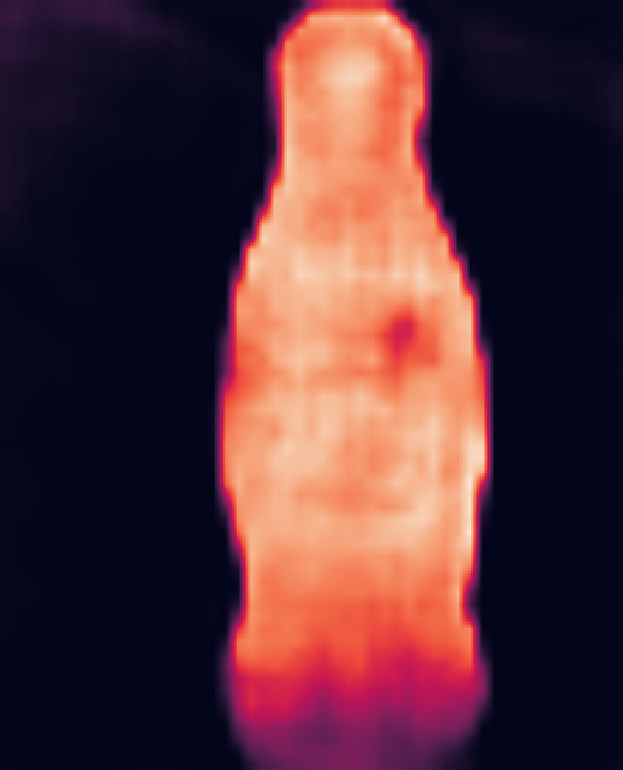}
	\end{subfigure}
	\begin{subfigure}[t]{0.135\textwidth}
		\includegraphics[width=1\textwidth, height=0.9\textwidth]{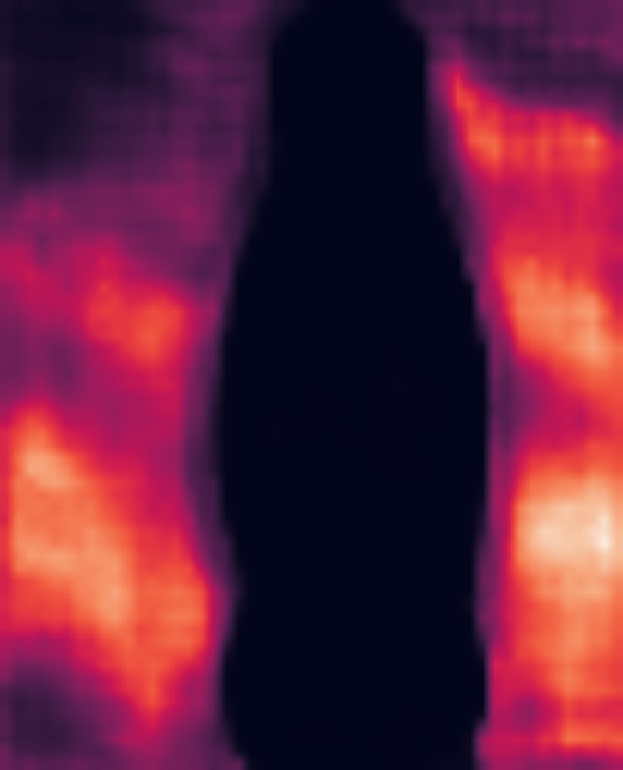}
	\end{subfigure}
	\begin{subfigure}[t]{0.135\textwidth}
		\includegraphics[width=1\textwidth, height=0.9\textwidth]{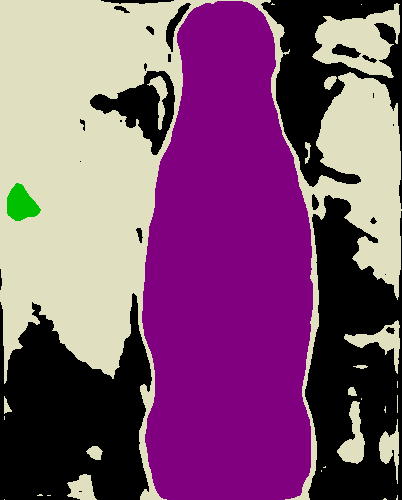}
	\end{subfigure}
	\begin{subfigure}[t]{0.135\textwidth}
		\includegraphics[width=1\textwidth, height=0.9\textwidth]{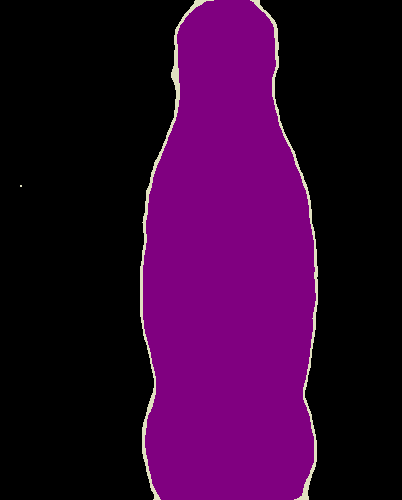}
	\end{subfigure}
 	\begin{subfigure}[t]{0.135\textwidth}
		\includegraphics[width=1\textwidth, height=0.9\textwidth]{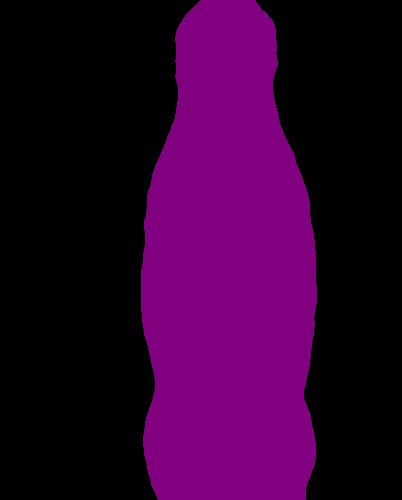}
	\end{subfigure}
 	\begin{subfigure}[t]{0.135\textwidth}
		\includegraphics[width=1\textwidth, height=0.9\textwidth]{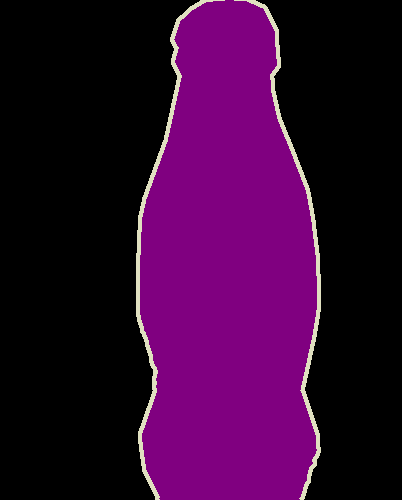}
	\end{subfigure}
    \begin{subfigure}[t]{0.135\textwidth}
		\includegraphics[width=1\textwidth, height=0.9\textwidth]{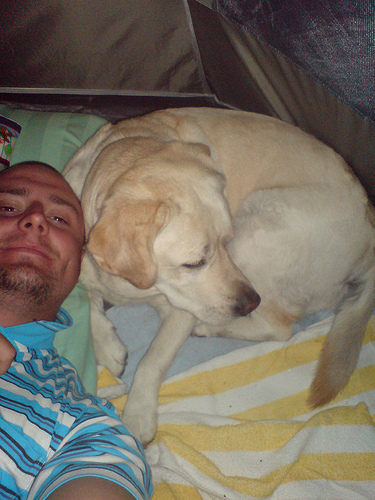}
	\end{subfigure}
	\begin{subfigure}[t]{0.135\textwidth}
		\includegraphics[width=1\textwidth, height=0.9\textwidth]{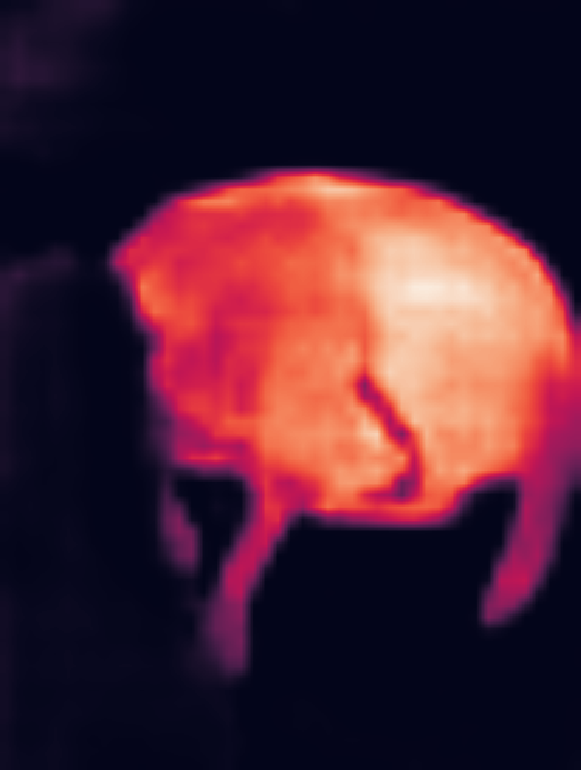}
	\end{subfigure}
	\begin{subfigure}[t]{0.135\textwidth}
		\includegraphics[width=1\textwidth, height=0.9\textwidth]{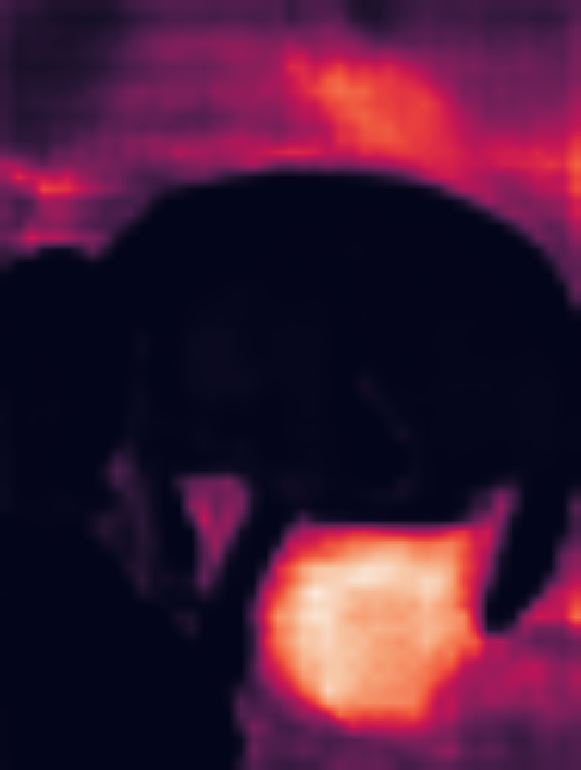}
	\end{subfigure}
	\begin{subfigure}[t]{0.135\textwidth}
		\includegraphics[width=1\textwidth, height=0.9\textwidth]{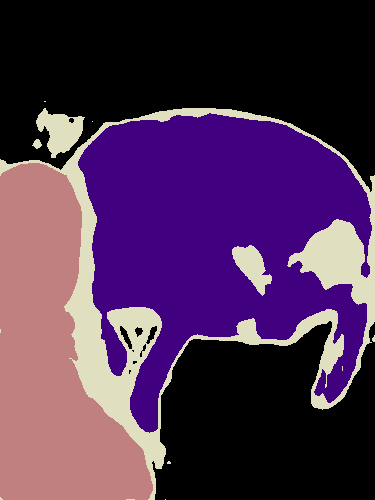}
	\end{subfigure}
	\begin{subfigure}[t]{0.135\textwidth}
		\includegraphics[width=1\textwidth, height=0.9\textwidth]{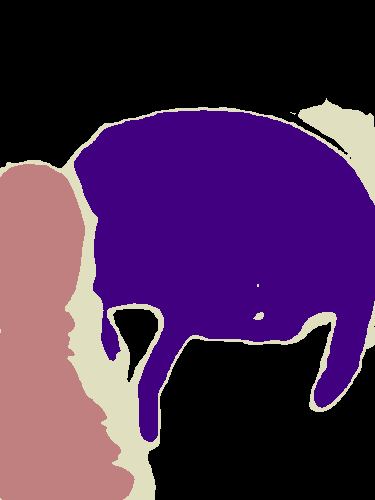}
	\end{subfigure}
 	\begin{subfigure}[t]{0.135\textwidth}
		\includegraphics[width=1\textwidth, height=0.9\textwidth]{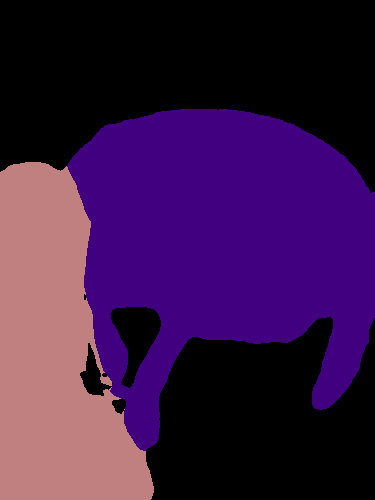}
	\end{subfigure}
 	\begin{subfigure}[t]{0.135\textwidth}
		\includegraphics[width=1\textwidth, height=0.9\textwidth]{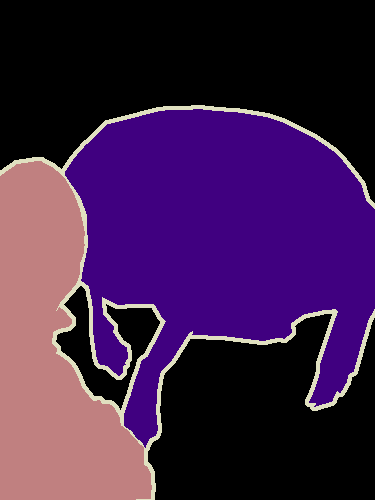}
	\end{subfigure}
    
  \begin{subfigure}[t]{0.135\textwidth}
		\includegraphics[width=1\textwidth, height=0.9\textwidth]{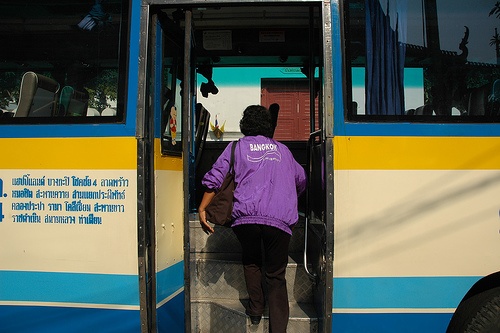}
	\end{subfigure}
	\begin{subfigure}[t]{0.135\textwidth}
		\includegraphics[width=1\textwidth, height=0.9\textwidth]{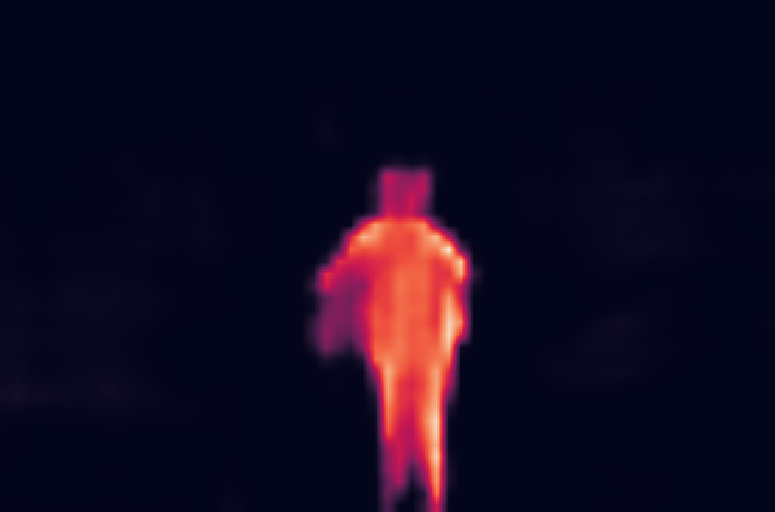}
	\end{subfigure}
	\begin{subfigure}[t]{0.135\textwidth}
		\includegraphics[width=1\textwidth, height=0.9\textwidth]{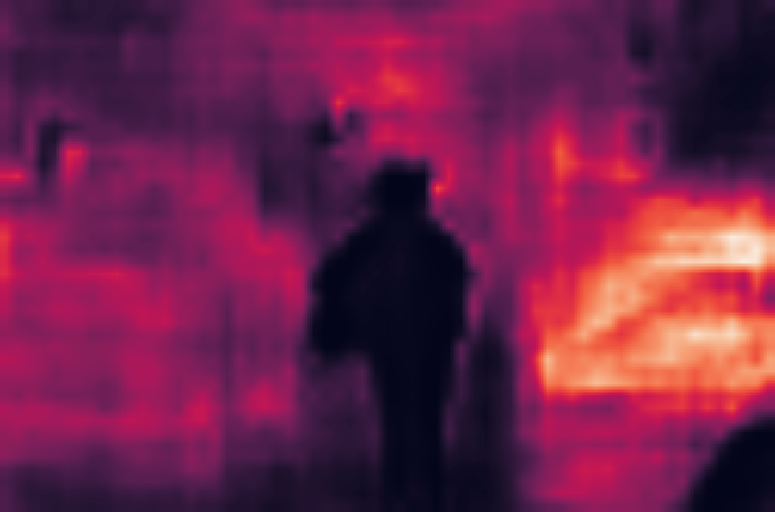}
	\end{subfigure}
	\begin{subfigure}[t]{0.135\textwidth}
		\includegraphics[width=1\textwidth, height=0.9\textwidth]{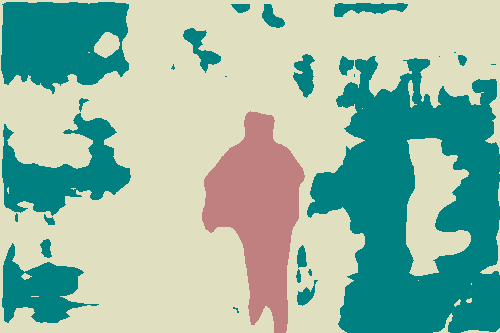}
	\end{subfigure}
	\begin{subfigure}[t]{0.135\textwidth}
		\includegraphics[width=1\textwidth, height=0.9\textwidth]{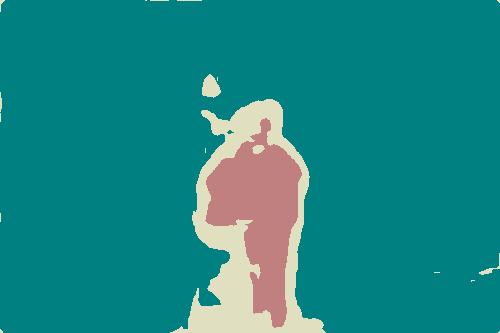}
	\end{subfigure}
 	\begin{subfigure}[t]{0.135\textwidth}
		\includegraphics[width=1\textwidth, height=0.9\textwidth]{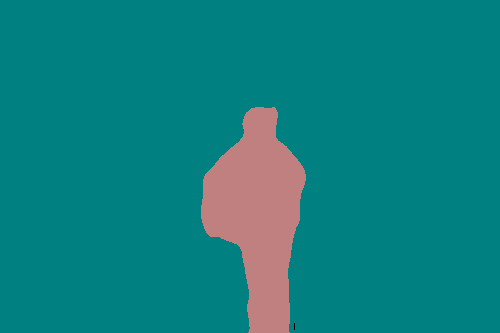}
	\end{subfigure}
 	\begin{subfigure}[t]{0.135\textwidth}
		\includegraphics[width=1\textwidth, height=0.9\textwidth]{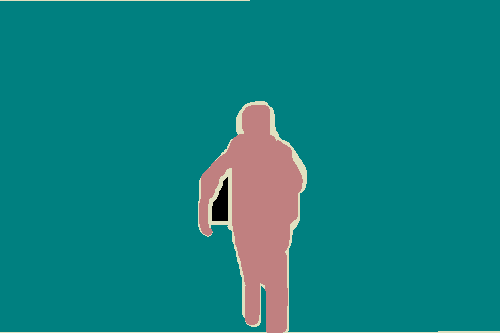}
	\end{subfigure}
      \begin{subfigure}[t]{0.135\textwidth}
		\includegraphics[width=1\textwidth, height=0.9\textwidth]{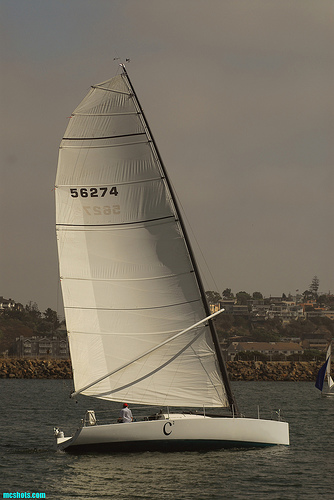}
	\end{subfigure}
	\begin{subfigure}[t]{0.135\textwidth}
		\includegraphics[width=1\textwidth, height=0.9\textwidth]{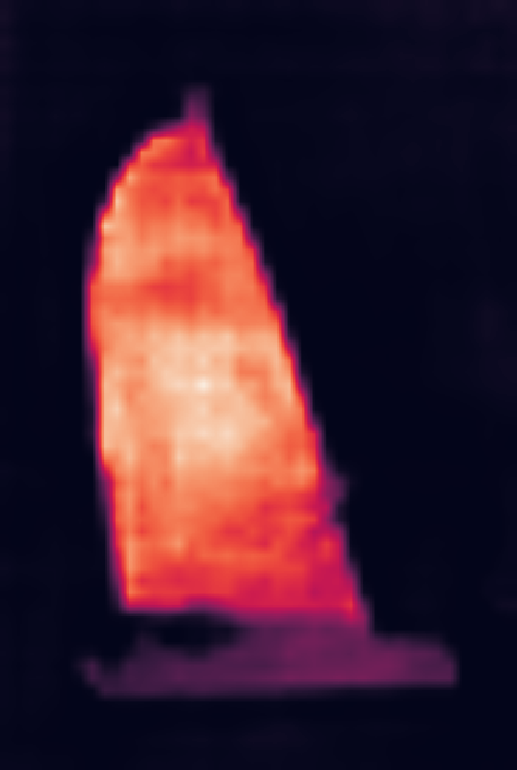}
	\end{subfigure}
	\begin{subfigure}[t]{0.135\textwidth}
		\includegraphics[width=1\textwidth, height=0.9\textwidth]{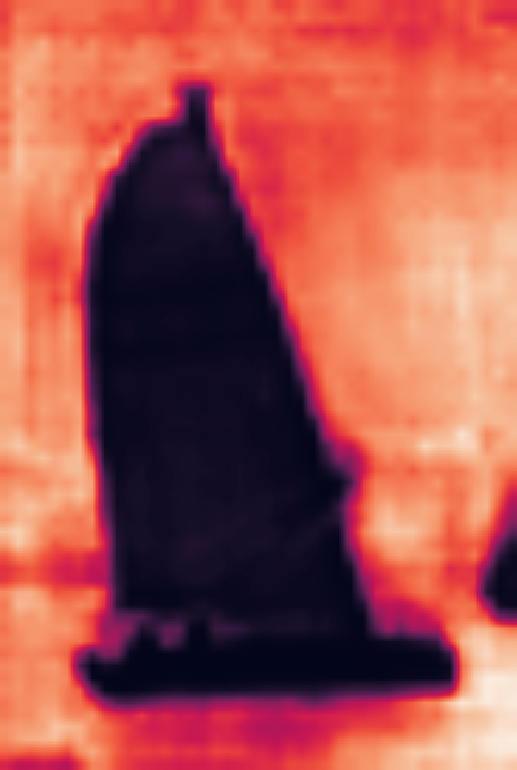}
	\end{subfigure}
	\begin{subfigure}[t]{0.135\textwidth}
		\includegraphics[width=1\textwidth, height=0.9\textwidth]{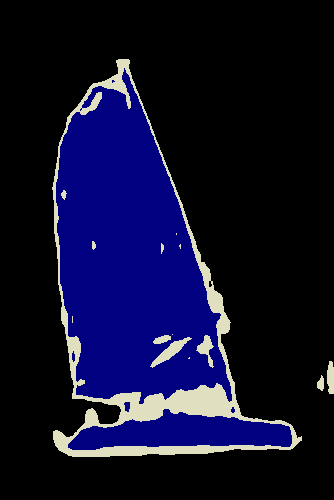}
	\end{subfigure}
	\begin{subfigure}[t]{0.135\textwidth}
		\includegraphics[width=1\textwidth, height=0.9\textwidth]{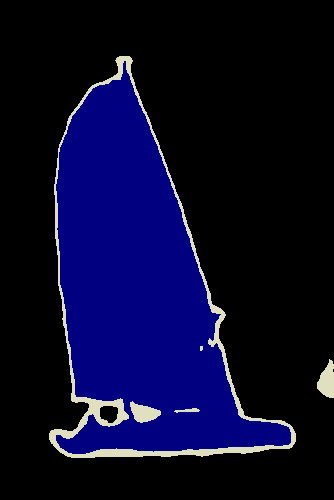}
	\end{subfigure}
 	\begin{subfigure}[t]{0.135\textwidth}
		\includegraphics[width=1\textwidth, height=0.9\textwidth]{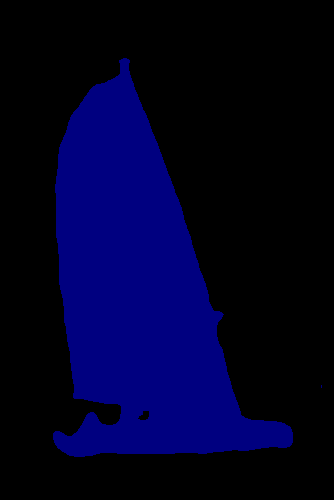}
	\end{subfigure}
 	\begin{subfigure}[t]{0.135\textwidth}
		\includegraphics[width=1\textwidth, height=0.9\textwidth]{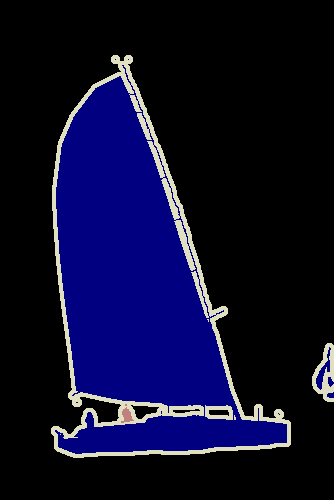}
	\end{subfigure}
   \begin{subfigure}[t]{0.135\textwidth}
		\includegraphics[width=1\textwidth, height=0.9\textwidth]{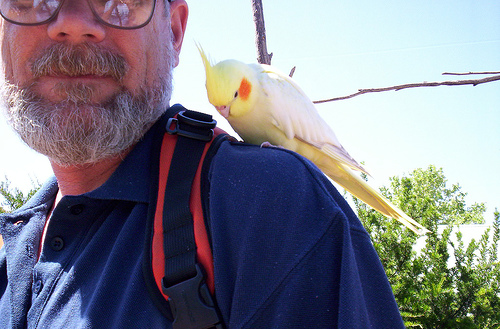}
	\end{subfigure}
	\begin{subfigure}[t]{0.135\textwidth}
		\includegraphics[width=1\textwidth, height=0.9\textwidth]{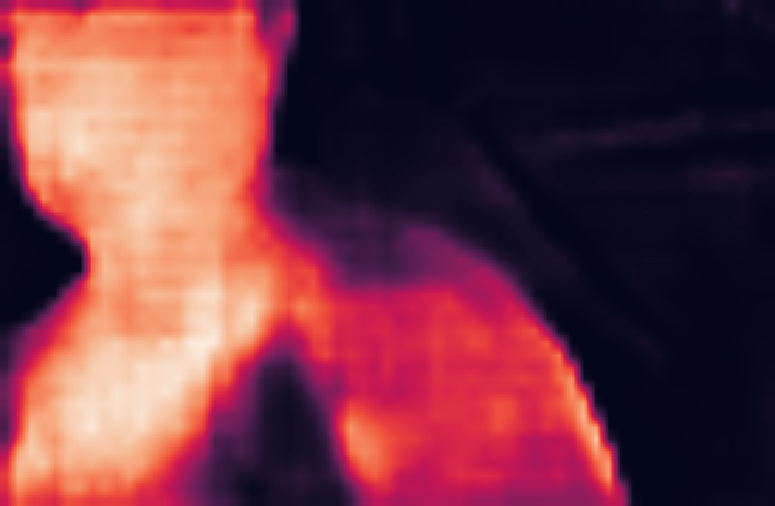}
	\end{subfigure}
	\begin{subfigure}[t]{0.135\textwidth}
		\includegraphics[width=1\textwidth, height=0.9\textwidth]{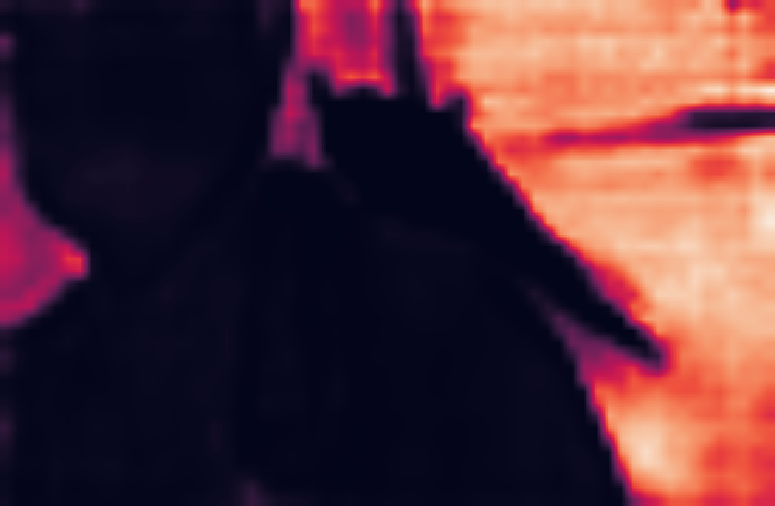}
	\end{subfigure}
	\begin{subfigure}[t]{0.135\textwidth}
		\includegraphics[width=1\textwidth, height=0.9\textwidth]{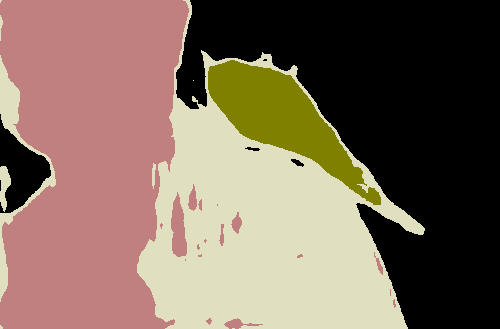}
	\end{subfigure}
	\begin{subfigure}[t]{0.135\textwidth}
		\includegraphics[width=1\textwidth, height=0.9\textwidth]{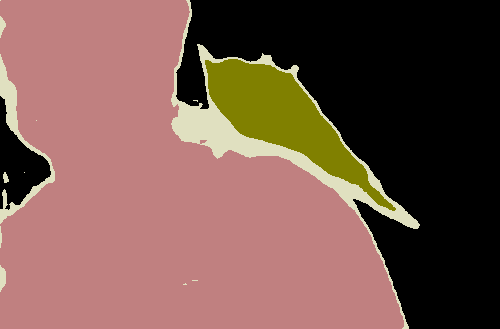}
	\end{subfigure}
 	\begin{subfigure}[t]{0.135\textwidth}
		\includegraphics[width=1\textwidth, height=0.9\textwidth]{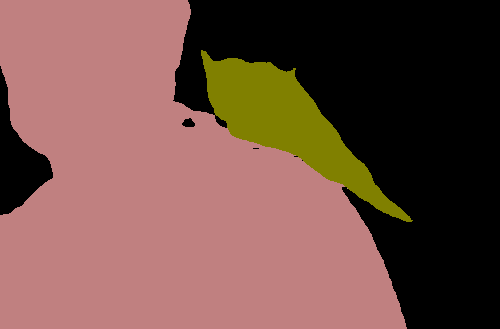}
	\end{subfigure}
 	\begin{subfigure}[t]{0.135\textwidth}
		\includegraphics[width=1\textwidth, height=0.9\textwidth]{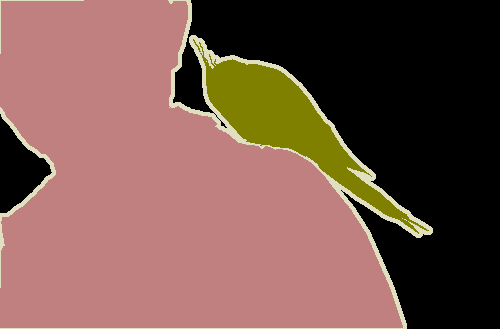}
	\end{subfigure}

    \begin{subfigure}[t]{0.135\textwidth}
		\includegraphics[width=1\textwidth, height=0.9\textwidth]{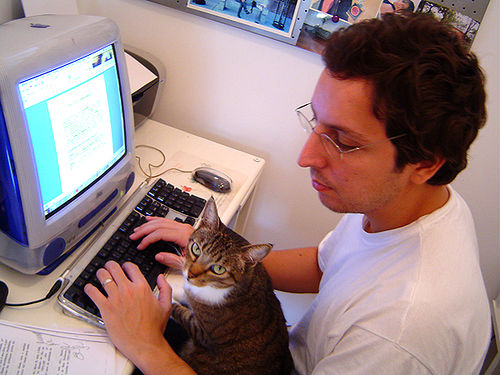}
        \subcaption{image}
	\end{subfigure}
	\begin{subfigure}[t]{0.135\textwidth}
		\includegraphics[width=1\textwidth, height=0.9\textwidth]{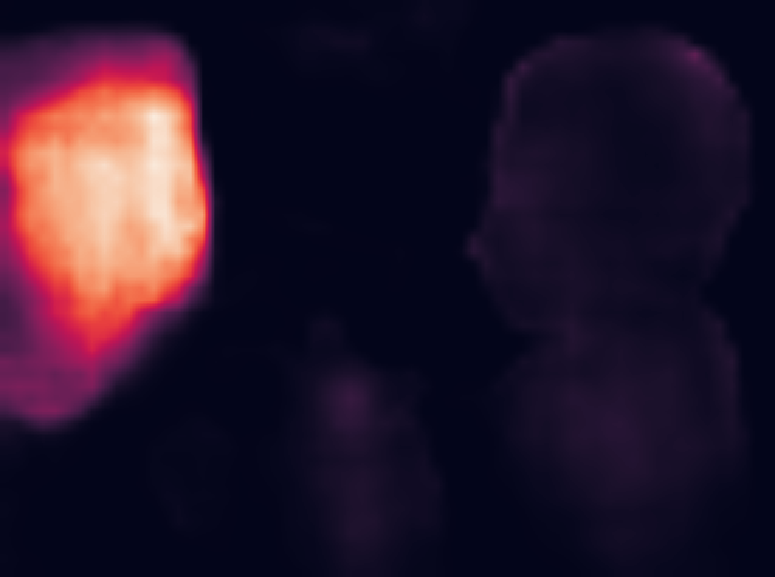}
        \subcaption{objects}
	\end{subfigure}
	\begin{subfigure}[t]{0.135\textwidth}
		\includegraphics[width=1\textwidth, height=0.9\textwidth]{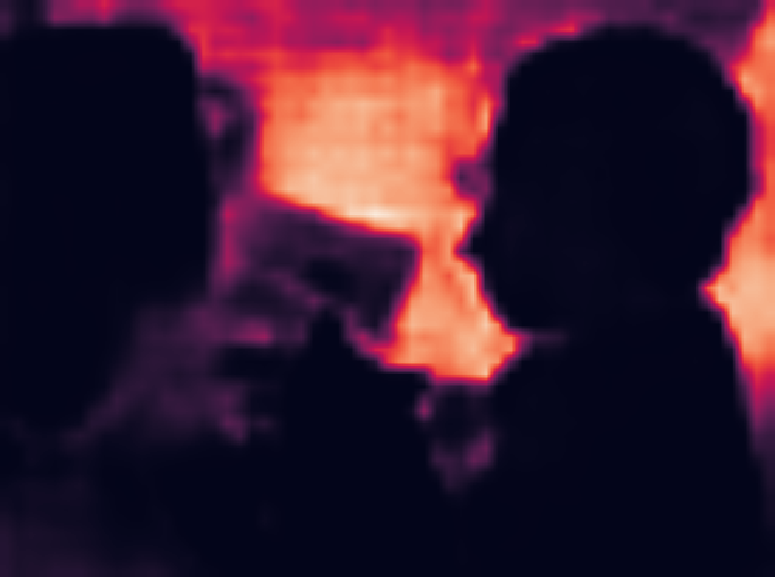}
        \subcaption{background}
	\end{subfigure}
	\begin{subfigure}[t]{0.135\textwidth}
		\includegraphics[width=1\textwidth, height=0.9\textwidth]{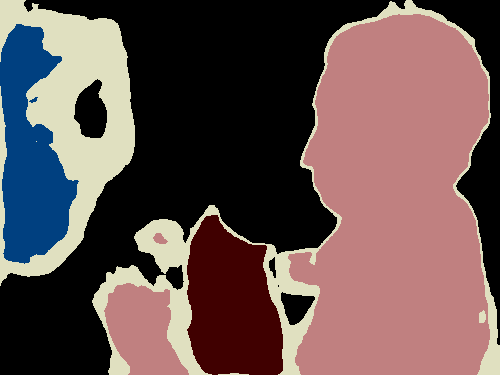}
        \subcaption{w/o corr}
	\end{subfigure}
	\begin{subfigure}[t]{0.135\textwidth}
		\includegraphics[width=1\textwidth, height=0.9\textwidth]{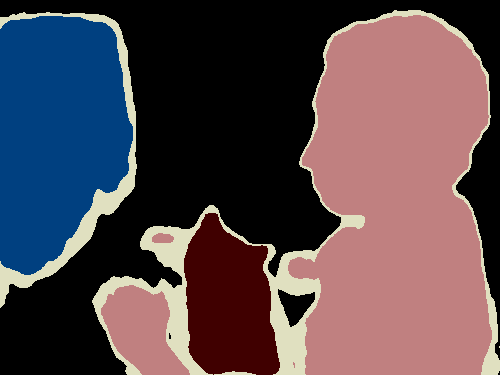}
        \subcaption{w/ corr}
	\end{subfigure}
 	\begin{subfigure}[t]{0.135\textwidth}
		\includegraphics[width=1\textwidth, height=0.9\textwidth]{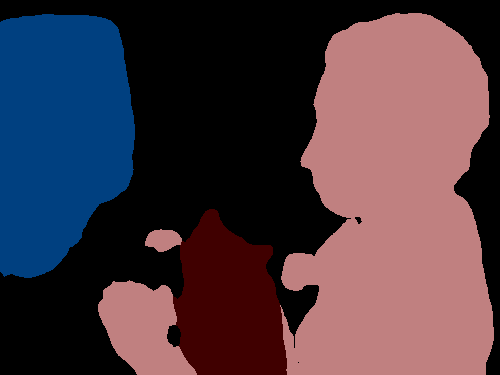}
        \subcaption{prediction}
	\end{subfigure}
 	\begin{subfigure}[t]{0.135\textwidth}
		\includegraphics[width=1\textwidth, height=0.9\textwidth]{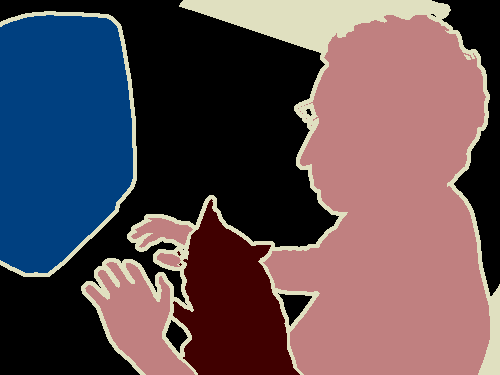}
        \subcaption{GT}
	\end{subfigure}
	\caption{More qualitative results from the val set of Pascal VOC 2012 dataset. (a) input image; (b) correlation map on object; (c) correlation map on background; (d) pseudo label without correlation matching; (e) pseudo label with \MyMthd{}; (f) prediction of \MyMthd{}; (g) ground truth. White areas in (d) and (e) are ignored regions due to low confidence.}
	\label{fig:more_pics}
\end{center}
\end{figure*}

In our main paper, we claim that proposed label propagation strategies can help mining reliable regions and we have verified this through both extensive quantitative and qualitative experiments. Here, we present more qualitative results in \figref{fig:more_pics} to further support our conclusion.

\section{More Analysis for Dynamic Threshold}
\subsection{Why Semi-supervised Semantic Segmentation Needs a Special Dynamic Threshold Design} \label{special design}
In this paper, besides the two label propagation strategies, we also propose a dynamic global threshold for semi-supervised semantic segmentation. Here we would like to discuss such an issue: \textbf{since the dynamic threshold strategy has been widely explored in many semi-supervised learning works, why does semi-supervised semantic segmentation need a special dynamic threshold design?}

Semi-supervised learning is different from the semi-supervised semantic segmentation task. 
We first present some potential differences between semi-supervised learning and semi-supervised semantic segmentation.

    \begin{table}[t]
    \centering
        \small
    \caption{ Comparison of \MyMthd{} with different thresholding strategies on PASCAL VOC 2012 val set with mIoU (\%) $\uparrow$ metric.}
    \label{tab: different thresholding}
\begin{tabular}{lcc} \toprule
 Method & 1 / 16(92)  & Full (1464) \\
\midrule 
CorrMatch & \textbf{76.4}  & \textbf{81.8}\\
Per-pixel thresholding& 64.1  & 77.2 \\
Update with maximum confidence& 63.4  & 74.4 \\
Update with average confidence & 75.4 & 80.2 \\
\bottomrule 
\end{tabular}
\end{table}
\begin{enumerate}
    \item \textbf{Task Objective:} In semi-supervised learning, the goal is to predict at the image level. In contrast, semi-supervised semantic segmentation is a dense prediction task and focuses on pixel-wise prediction. Its objective is to classify each pixel individually and there might be multiple classes presented in an image.
    \item \textbf{Threshold Usage:} For semi-supervised learning, the threshold is typically applied to determine whether the prediction of an image is regarded as the pseudo label. Meanwhile, for semi-supervised semantic segmentation, the threshold is applied to individual pixels to screen high-confidence regions and treat them as pseudo labels.
    \item \textbf{Object Size:} For semi-supervised learning, the model is trained to classify the input image. However, semi-supervised semantic segmentation aims to segment the image into distinct regions for different semantic objects. Since objects in an image often have diverse sizes, and their corresponding feature distributions may vary significantly, the learning difficulties tend to be various.
\end{enumerate}

Taking the above potential differences into account, in \tabref{tab: different thresholding}, we conduct some corresponding experiments to demonstrate that simply extending the strategies of semi-supervised learning into a pixel-wise paradigm is not sufficient enough and our design for semi-supervised semantic segmentation is non-trivial. 
\begin{enumerate}
    \item \textbf{Per-pixel thresholding:} Firstly, we set a threshold for each pixel and update them with corresponding confidence individually. However, since the positions of objects with different semantics are not fixed, and their confidence distribution is not determined by the pixel position, this scheme has obvious performance degradation.
    \item \textbf{Update with maximum confidence:} Then, we conduct experiments by using a global threshold for each class and updating the threshold with global maximum confidence. However, some pixels are easier to learn and exhibit confidence values very close to 1. This makes the threshold quickly close to 1, causing most regions treated as low-confidence ones. The performance drops.
    \item \textbf{Update with average confidence:} Finally, we conduct experiments using the global average values to update the threshold, and each pixel participated in the threshold update equally. However, different classes often occupy different numbers of pixels and have different learning difficulties. For instance, background often occupies a large part of the images and tends to maintain high confidence in the Pascal VOC 2012 datasets. Thus, this scheme still introduces a relatively high threshold and causes performance degradation.
\end{enumerate}

Overall, different from semi-supervised learning, designing a strategy for semi-supervised semantic segmentation requires the consideration of spatial dependencies and pixel-wise predictions, making it more complex and challenging. Our strategy takes the aforementioned differences into account by considering the maximum confidence for each class appearing in the predictions and employs their average value to maintain the dynamic threshold. Experimental results show that our threshold update strategy is non-trivial. Furthermore, to our knowledge, we are the first to introduce a dynamic threshold and label propagation into semi-supervised semantic segmentation.

\subsection{Why not Per-class Threshold Updating}
Considering that the proposed threshold strategy is updating a global threshold after all, it might be argued that using a dynamic threshold updating strategy for each class may lead to performance improvements since it has shown success in semi-supervised classification tasks~\cite{wang2022freematch, zhang2021flexmatch}. 
However, as discussed in \secref{special design}, the classification and semantic segmentation tasks have different characteristics.
%
%
Therefore, a similar strategy may be not suitable for semi-supervised semantic segmentation tasks. 
To further illustrate this point, we conduct the following per-class thresholding strategy.

We first initialize a tensor with the same size as the number of categories, and its value is the same as the global initialization value. We use a similar EMA style to iteratively update strategy as global threshold updating. For each predicted class $l$ in model predictions $\mathcal{F}(x^w_i)$, the process for each iteration is defined as:
\begin{equation}
\tau'_l = \max [ \mathbbm{1}(\mathcal{F}(x^w_i) = l) \circ \max^{c}(\hat{\mathcal{F}}(x^w_i))],
\end{equation}
where $\hat{\mathcal{F}}(x^w_i)$ is the logits prediction of unlabeled images with weak data augmentations. 
This operation means we take the maximum confidence of each predicted class in weakly augmented unlabeled images and consider them as the increment for each class at each iteration. Then, similar to FreeMatch~\cite{wang2022freematch}, we use maximum normalization operation to integrate the global and local thresholds.

We conduct experiments on the original Pascal VOC 2012 dataset with 321$\times$321 training size in \tabref{tab: per-class}. It can be clearly seen that converting it to a per-class scheme brings around 1\% performance drop compared to the global threshold updating strategy.

 \begin{table}[t]
    \centering
        \small
    \setlength{\tabcolsep}{0.5pt}
    \caption{ Comparison of \MyMthd{} with and without per-class thresholding strategy on PASCAL VOC 2012 val set with mIoU (\%) $\uparrow$ metric. $^*$ means with per-class threshold updating strategy. }
    \label{tab: per-class}
\begin{tabular}{lccccc} \toprule
 Method  & 1 / 16(92) & 1 / 8(183) &1 / 4(366) & 1 / 2(732) & Full (1464) \\
\midrule \MyMthd{}& \textbf{76.4} & \textbf{78.5}  & \textbf{79.4} &\textbf{ 80.6}  & \textbf{81.8} \\
\MyMthd{}$^*$ & 75.1 & 76.7  & 78.3 & 79.3  & 80.3\\
\bottomrule 
\end{tabular}
\end{table}

\newpage
{
\small
\bibliographystyle{ieee_fullname}
\bibliography{egbib}
}

\end{document}